\documentclass[lettersize,journal]{IEEEtran}
\usepackage[caption=false,font=normalsize,labelfont=sf,textfont=sf]{subfig}
\usepackage{cite}

\usepackage{hyperref}       
\usepackage{url}            
\usepackage{booktabs,siunitx}       
\usepackage{amsfonts}       
\usepackage{nicefrac}       
\usepackage{microtype}      
\usepackage{cite}
\usepackage{amsmath,amssymb,amsfonts}
\usepackage{algorithmic}
\usepackage{graphicx}
\usepackage{textcomp}
\usepackage{float}
\usepackage{array,multirow,graphicx}
\usepackage{animate}
\usepackage{arydshln}
\usepackage{mathtools}
\usepackage{xmpmulti}
\usepackage{alltt}
\usepackage{makecell}
\usepackage{bm}
\usepackage[dvipsnames]{xcolor}
\definecolor{ForestGreen}{RGB}{11, 148, 43}
\definecolor{magenta}{RGB}{247,29,176}
\definecolor{orange}{RGB}{219,106,72}
\definecolor{skyblue}{RGB}{42,138,153}
\definecolor{purple}{RGB}{153,51,255}
\usepackage{easybmat}
\usepackage{soul}
\usepackage{enumitem}
\usepackage{algorithmic}

\makeatletter
\renewcommand*{\@fnsymbol}[1]{\@alph{#1}}
\makeatother

\usepackage{xcolor}
\usepackage{dblfloatfix}

\DeclareMathOperator*{\argmin}{arg\,min}

\newcommand{\vect}[1]{\boldsymbol{\mathbf{#1}}}

\makeatletter
\renewcommand*{\@fnsymbol}[1]{\@alph{#1}}
\makeatother

\usepackage{algorithm}

\usepackage{eqparbox,collcell}
\usepackage{dsfont}

\newcommand{\columnalign}{}
\newcommand{\columnalignment}[2]{\renewcommand{\columnalign}{\eqmakebox[#1][#2]}}

\DeclareMathOperator{\bTheta}{\Theta}

\usepackage{amsfonts}

\usepackage{thmtools}
\usepackage{thm-restate}
\usepackage{cleveref}

\declaretheorem{assumption}
\declaretheorem{theorem}
\declaretheorem{lemma}

\newcommand{\qed}{\hfill \blacksquare}
\newcommand{\hide}[1]{}

\begin{document}

\title{Fast and Accurate Graph Learning for Huge Data via Minipatch Ensembles}

\author{Tianyi Yao, Minjie Wang, and Genevera I. Allen
\thanks{T. Yao (tianyi.bourne.yao@gmail.com) was in the Department of Statistics, Rice University, Houston, TX}
\thanks{M. Wang (wangmj@umn.edu) is in the School of Statistics, University of Minnesota, Minneapolis, MN}
\thanks{G. I. Allen (gallen@rice.edu) is in the Departments of Electrical and Computer Engineering, Statistics, and Computer Science, Rice University, Houston, TX.}
}

\maketitle

\begin{abstract}
Gaussian graphical models provide a powerful framework for uncovering conditional dependence relationships between sets of nodes; they have found applications in a wide variety of fields including sensor and communication networks, image processing and computer vision, physics, finance, and computational biology. Often, one observes data on the nodes and the task is to learn the graph structure, or perform graphical model selection.  While this is a well-studied problem with many popular techniques, there are typically three major practical challenges: i) many existing algorithms become computationally intractable in huge-data settings with tens of thousands of nodes; ii) the need for separate data-driven hyperparameter tuning considerably adds to the computational burden; iii) the statistical accuracy of selected edges often deteriorates as the dimension and/or the complexity of the underlying graph structures increase. We tackle these problems by developing the novel Minipatch Graph (MPGraph) estimator. Our approach breaks up the huge graph learning problem into many smaller problems by creating an ensemble of tiny random subsets of both the observations and the nodes, termed minipatches.  We then leverage recent advances that use hard thresholding to solve the latent variable graphical model problem to consistently learn the graph on each minipatch.  Our approach is computationally fast, embarrassingly parallelizable, memory efficient, and has integrated stability-based hyperparamter tuning.  Additionally, we prove that under weaker assumptions than that of the Graphical Lasso, our MPGraph estimator achieves graph selection consistency. We compare our approach to state-of-the-art computational approaches for Gaussian graphical model selection including the BigQUIC algorithm, and empirically demonstrate that our approach is not only more statistically accurate but also extensively faster for huge graph learning problems.
\end{abstract}

\begin{IEEEkeywords}
Graphical models, graph topology inference, graph structural learning, graphical model selection, Gaussian graphical models, graphical lasso, minipatch ensemble learning.
\end{IEEEkeywords}

\section{Introduction}
\label{sec:mpgraph-intro}

\IEEEPARstart{G}{aussian} graphical models are popular multivariate probability distributions for studying and uncovering conditional dependence relationships between variables or nodes \cite{lauritzen1996graphical}. Graphical models have found wide application in systems biology \cite{Sinoquet2013}, computer vision and image processing \cite{JI2020191}, communication and sensor networks \cite{doi:10.1155/2011/928958}, natural language processing \cite{mihalcea_radev_2011}, physics \cite{LEIFER20081899}, and finance \cite{Seregina2022}, among many others.  In several of these domains, we do not know the inherent graph structure and seek to learn this from data, a problem called graph structural learning or graphical model selection \cite{drton2017structure}.  With larger and larger data sets in a variety of domains, we are interested in performing graphical model selection on data with a huge number of nodes.  Consider an example from neuroimaging where Gaussian graphical models are often used to learn the functional connectivity between brain regions or neurons \cite{friston2011functional,yatsenko2015improved,chang2019graphical}.  In fMRI data, the number of brain regions and hence graph nodes can be on the order of tens- to hundreds-of-thousands \cite{friston2011functional} whereas in calcium imaging, the number of neurons or graph nodes in some newer technologies can measure in the tens-of-thousands \cite{stringer2019computational,vinci2019graph}.

Many existing Gaussian graphical model selection techniques are computationally too cumbersome to be applied in such large data settings \cite{hsieh2013bigquic}.  Further, these techniques require hyperparamter tuning to determine the sparsity or edge set of the graph; this process often creates an insurmountable computational burden for large data.  Finally, the statistical accuracy of Gaussian graphical model selection methods is known to degrade with a large number of nodes, creating further challenges \cite{drton2017structure}.  In this paper, our goal is to develop a radically different type of computational method for learning the structure of Gaussian graphical models that not only yields statistically more accurate estimation, but is also more scalable for huge data.

Formally, let $X=(X_1, X_2,\hdots,X_M)$ be a $M$-dimensional random vector following a multivariate Gaussian distribution $\mathcal{N}(\mathbf 0, \vect{\Sigma})$ with covariance matrix $\vect{\Sigma}$ and inverse covariance matrix $\vect{\Theta} = \vect{\Sigma}^{-1}$. Graphical models are often denoted by $G=(V, E)$, in which the node set $V$ represents the collection of random variables $X$ and edge set $E$ characterizes the conditional dependence relationship between these random variables. An absence of edge between $X_i$ and $X_j$ means that $X_i$ and $X_j$ are conditionally independent; for Gaussian graphical models, this corresponds to zero entries in the precision matrix $\Theta_{ij}=\Theta_{ji}=0$. One important objective of Gaussian graphical model selection or learning the graph structure is to infer the edge set $E$, or the sparsity pattern of $\vect{\Theta}$, from the observed data.

\subsection{Related Works}
\label{sec:mpgraph-literature}

One popular approach to Gaussian graphical model selection problem is the sparse inverse covariance estimator introduced by \cite{yuan2007model, banerjee2008model}. Since its inception, a plethora of optimization algorithms \cite{duchi2008graph,d2008first,rolfs2012ista} have been proposed to solve this problem, including the widely-used graphical Lasso procedure \cite{friedman2008sparse}. However, as noted in \cite{hsieh2013bigquic}, many of these solvers do not scale to problems with thousands or tens of thousands of nodes, which is commonly observed in modern applications. The BigQUIC algorithm \cite{hsieh2013bigquic} is arguably the state-of-the-art computational approach to the sparse inverse covariance estimation. While numerical studies \cite{hsieh2013bigquic} have shown that it can solve the estimation problem with a million nodes within a day, these experiments were conducted in somewhat ideal conditions where the single optimal tuning hyperparameter value is assumed to be known a priori, which is rare in practice. The statistical and computational performance of the BigQUIC algorithm in more realistic settings with data-driven hyperparameter tuning have been inadequately investigated.

A related line of work seeks to lower the computational cost of estimating sparse Gaussian graphical models by soft-thresholding the sample covariance matrix prior to directly solving the sparse inverse covariance estimation problem \cite{witten2011new,mazumder2012exact,fattahi2019graphical, zhang2020fst}. However, these approaches generally require imposing specific sparsity constraints such as assuming the underlying graph has connected components or a chordal structure, which are often too restrictive to hold in real-world applications of Gaussian graphical model selection.

In addition, \cite{meinshausen2006high} proposed the neighborhood selection procedure, which separately fits the Lasso regression estimator to learn the neighborhood of each node and then infers the edge set by aggregating the neighbors selected from all nodewise regressions. \cite{liu2017tiger} further extended the neighborhood selection approach by replacing the Lasso estimator with the more tuning-insensitive SQRT-Lasso estimator \cite{belloni2011sqrtlasso}, which yields the TIGER method for estimating high-dimensional Gaussian graphical models. Another line of work proposed to employ the Dantzig selector for solving each nodewise sparse regression subproblem, resulting in the CLIME method \cite{cai2011constrained}. Despite generally faster computation due to the easily parallelizable nature of the nodewise regressions, these Gaussian graphical model selection strategies can still become computationally challenging as the number of nodes $M$ increases because they need to solve $M$-dimensional sparse regression problems for a total of $M$ times even for a single value of tuning hyperparameter.

Importantly, most of the aforementioned computational approaches to the Gaussian graphical model selection problem have at least one tuning hyperparameter that controls how many edges and which edges are selected, thus significantly impacting the statistical accuracy of the estimated sets of edges \cite{muller2016generalized}. Even though many of the existing methods were shown to enjoy certain guarantees for the accuracy of the edges that they select, such theoretical guarantees are often contingent on some oracle tuning hyperparameter choices that cannot be used in practice, as astutely pointed out by \cite{liu2017tiger}. In realistic applications of Gaussian graphical model selection, these existing approaches need to be employed in conjunction with data-driven hyperparameter selection procedures such as cross-validation \cite{efron1982jackknife, wasserman2009high}, extended BIC \cite{chen2008ebic, foygel2010extended}, StARS \cite{liu2010stability}, and B-StARS \cite{muller2016generalized}. However, the statistical accuracy of selected edges often degrades as the dimension and/or the complexity of the underlying graph structures increase when the tuning hyperparameters are chosen via these data-dependent ways without oracle tuning information. Additionally, these data-driven hyperparameter tuning strategies require solving the graph selection problem at least once for each candidate tuning hyperparameter value, thus considerably adding to the overall computational burden.

{\bf Contributions:} We summarize our main contributions as follows: We develop a statistically accurate and fast computational approach to learn the structure of Gaussian graphical models named the Minipatch Graph (MPGraph) estimator that ensembles thresholded graph estimators trained on tiny, random subsets of both observations and nodes, termed minipatches (see Sec.~\ref{sec:mpgraph-method}). Leveraging such ensembles, our approach not only selects edges with improved statistical accuracy, but is also computationally fast with integrated stability-based hyperparameter tuning. Additionally, we theoretically analyze MPGraph showing that it attains graph selection consistency under weaker assumptions than many existing graph estimators. We further empirically demonstrate the practical advantages of our approach through extensive comparative studies, showing that the MPGraph estimator dominates state-of-the-art computational approaches to Gaussian graphical model selection in terms of both edge selection accuracy and computational time for huge graph selection problems.  The major innovation of our work lies in the idea to split up huge graph selection problems into smaller and more computationally manageable units (minipatches) and then ensemble the results back together to obtain a consistent graph estimate. An implementation of our proposed method is provided in the open-source \texttt{Python} package \texttt{minipatch-learning}, available at \href{https://github.com/DataSlingers/minipatch-learning}{https://github.com/DataSlingers/minipatch-learning}.

\section{Minipatch Graph Learning}
\label{sec:mpgraph-method}

\subsection{Review: Minipatch Learning}

The idea of using random subsets of observations and/or features for model training has appeared in many areas of machine learning including random forests \cite{breiman2001}, stochastic optimization \cite{hardt2016sgd}, and dropout training \cite{JMLR:v15:srivastava14a}. Recently, a line of work in feature selection \cite{yao2021feature} and ensemble learning \cite{yao2021minipatch,toghani2021mpboost} coined the term ``minipatches'' to denote tiny, random subsets of both observations and features of the data. Following the notations in \cite{yao2021minipatch}, given observed data matrix $\mathbf{X}\in\mathbb{R}^{N\times M}$ that comprises $N$ observations each having $M$ features, a minipatch is obtained by simultaneously subsampling $n$ rows (observations) and $m$ columns (features) without replacement using some form of randomization, usually with $n\ll N$ and $m\ll M$.  Notice that random minipatch ensemble learning is embarrassingly parallelizable, memory efficient, and learning on each tiny minipatch is notably faster, offering major computational advantages.

\subsection{Minipatch Graph Algorithm (MPGraph)}
\label{sec:mpgraph-mpgraph}

We propose a new approach to graph learning that leverages minipatch ensembles to tackle both the statistical and computational challenges with Gaussian graphical model selection in huge-data settings.  At first glance, directly applying the idea of minipatch learning to graphical model selection seems straightforward as one might ask whether we can simply train separate graph estimators on each minipatch and then aggregate all these subgraph estimates to form a complete graph. However, such simplistic approach is actually problematic due to the complex dependency structure between the nodes/variables in the data. Concretely, suppose that we take a random minipatch that consists of a subset of nodes from the original data set and try to estimate the conditional dependence graph between these nodes by fitting a graph estimator to this minipatch. Then, the remaining nodes outside of this minipatch effectively become latent unobserved nodes with respect to the minipatch, thus inducing many false positive edges of small magnitude in this subgraph estimate \cite{vinci2019graph}.  Estimating graphs for each minipatch thus leads to the well-known and challenging latent variable graphical model problem \cite{chandrasekaran2012}.  \cite{chandrasekaran2012} propose to solve this problem by assuming that the latent variables induce a low-rank effect on the observed nodes and propose a sparse and low-rank graphical model estimator.  Unfortunately, it is easy to see that if we take $m \ll M$ in our minipatch learning framework, which means we have many more latent nodes than observed nodes, and seek to learn general graph structures, then this low-rank assumption is violated.

At this point, it seems like we might not be able to take advantage of minipatch learning as we have substituted the challenge of large-scale graph learning for an equally challenging problem of dealing with many latent nodes in the context of graph learning.  But if it were possible to quickly solve the latent variable graphical model problem to yield a consistent graph structural estimate for each minipatch, then we could ensemble the subgraph estimates and fully realize the potential of minipatch learning. To achieve this, we propose to utilize a simple and computationally fast approach recently proposed by \cite{wang2021tglasso} to solve each latent variable problem. They apply a hard thresholding operator to the graphical Lasso (GLasso) estimate and show that this is sparsistent for Gaussian graphical model selection in the presence of latent variables under fairly weak assumptions. Importantly, their approach does not assume low-rankness and is thus applicable in the $m \ll M$ setting.  Putting everything together, we propose a new graph structural learning algorithm, MPGraph, which utilizes minipatch learning and thresholded graph estimation techniques to select statistically accurate edges in a computationally efficient manner.

{\centering
\setcounter{algorithm}{0}
\begin{algorithm}[!ht]
   \caption{MPGraph Algorithm}
   \label{alg:mpgraph}
\begin{algorithmic}
   \STATE {\bfseries Input:} $\mathbf{X}\in\mathbb{R}^{N\times M}$, $n$, $m$, $\pi_{\text{thr}}\in(0,1)$
   \vspace{2pt}
   \FOR[ In parallel]{$k=1,2,\hdots,K$}
   \vspace{4pt}
   \STATE 1) Sample a minipatch: subsample $n$ observations $I_k\subset [N]$ and $m$ features $F_k\subset [M]$ uniformly at random without replacement to get a minipatch $\mathbf{X}_{I_k,F_k}\in\mathbb{R}^{n\times m}$
   \vspace{4pt}
   \STATE 2) Fit base thresholded graph estimator to $\mathbf{X}_{I_k,F_k}$ to obtain estimated subgraph $\widetilde{\vect{\Theta}}^{(k)}\in\mathbb{R}^{m\times m}$
   \vspace{4pt}
   \STATE 3) Map indices: for each pair of nodes in the minipatch $\{(i,j)\in F_k\times F_k: i < j\}$, set $(i', j')=\{(a,b)\in [m]\times [m]: F_{k,a}=i, F_{k,b}=j\}$
   \vspace{4pt}
   \STATE 4) Update edge selection indicator $S_{ij}^{(k)}$ and node sampling indicator $D_{ij}^{(k)}$, $\forall\: 1\leq i < j \leq M$:
   \begin{align*}
       S_{ij}^{(k)} &= \mathds{1}(i\in F_k, j\in F_k, \widetilde{\vect{\Theta}}^{(k)}_{i',j'}\neq 0) \\ D_{ij}^{(k)} &= \mathds{1}(i\in F_k, j\in F_k)
   \end{align*}
   \ENDFOR
   \STATE Compute edge selection frequencies $\hat{\Pi}_{ij}^{(K)}, \forall\: 1\leq i < j \leq M$:
    \begin{align*}
        \hat{\Pi}^{(K)}_{ij} = \frac{\sum_{k=1}^K S_{ij}^{(k)}}{\text{max}(1,\sum_{k=1}^K D_{ij}^{(k)})}
    \end{align*}
   \STATE {\bfseries Output:} $\hat{E}^{\text{stable}} = \big\{1\leq i < j \leq M: \hat{\Pi}^{(K)}_{ij} \geq \pi_{\text{thr}} \big\}$
\end{algorithmic}
\end{algorithm}
}

Our proposed MPGraph algorithm is summarized in Algorithm \ref{alg:mpgraph}. Following the notational conventions, $\mathbf{X}_{I_k,F_k}$ denotes the submatrix of $\mathbf{X}$ containing its rows indexed by $I_k$ and its columns indexed by $F_k$. Similarly, $\vect{\Theta}_{i,j}$ denotes the element at the $i$th row and $j$th column. Also, $F_{k,a}$ represents the $a$th element in the vector $F_k$. For simplicity, $[N]$ denotes the set $\{1,2,\hdots,N\}$. MPGraph fits base thresholded graph estimators to many tiny, random minipatches. On the $k$th minipatch, the node sampling indicator $D^{(k)}_{ij}$ is set to 1 if and only if both nodes $i$ and $j$ are sampled into this minipatch; similarly, the edge selection indicator $S^{(k)}_{ij}$ is set to 1 if there is an edge between node $i$ and $j$, as determined by the base thresholded graph estimator. After $K$ iterations, MPGraph computes the edge selection frequencies by taking an ensemble of edge selection events over all minipatches. We define the selection frequency of the edge between node $i$ and $j$, $\hat{\Pi}^{(K)}_{ij}$, to be the number of times both nodes are sampled together and have an estimated edge between them as determined by the base graph estimator divided by the number of times both are sampled together into minipatches. MPGraph finally produces a set of stable edges $\hat{E}^{\text{stable}}$ whose selection frequencies are above a user-specific threshold $\pi_{\text{thr}}\in(0,1)$. We discuss the choice of $\pi_{\text{thr}}$ in Sec.~\ref{sec:param}. Because the computations on each minipatch can be done independently, the MPGraph algorithm is embarrassingly parallelizable, which can yield enormous computational savings.

As mentioned, we adopt the hard thresholded GLasso (TGLasso) approach from \cite{wang2021tglasso} as the latent variable graphical model estimator in Step 2 of Algorithm~\ref{alg:mpgraph}. We describe this in detail in Algorithm~\ref{alg:thresholdedglasso}. We save the investigations of other types of internal graph estimation strategies for future work. Given a minipatch, Algorithm \ref{alg:thresholdedglasso} fits the GLasso at a small amount of regularization $\lambda_0$ to get an initial subgraph estimate $\hat{\vect{\Theta}}^{\lambda_0}_{k}$, as suggested by \cite{wang2021tglasso}. After that, it hard-thresholds the initial estimate at a sequence of threshold values $\tau_k$ and eventually outputs the best thresholded subgraph estimate $\widetilde{\vect{\Theta}}^{(k)}$ according to the eBIC criterion. $\gamma$ is a user-specific parameter for controlling the amount of regularization from the eBIC criterion. We found setting $\gamma$ to a default value of $0.5$ works well in practice.

{\centering
\setcounter{algorithm}{1}
\begin{algorithm}[H]
   \caption{Thresholded Graph Estimator for Step 2) in Algorithm \ref{alg:mpgraph}}
   \label{alg:thresholdedglasso}
\begin{algorithmic}
   \STATE {\bfseries Input:} $\mathbf{X}_{I_k,F_k}$, $\gamma\geq 0$, $\omega_0>0$
   \vspace{4pt}
   \STATE 1) Fit GLasso to minipatch $\mathbf{X}_{I_k,F_k}$ at regularization parameter $\lambda_0=\omega_0\sqrt{\log{m} /n}$ to get $\hat{\vect{\Theta}}^{\lambda_0}_{k}$
   \vspace{4pt}
   \STATE 2) Extract the maximum off-diagonal element: $$\theta_k=\max_{1\leq i<j\leq m}|[\hat{\vect{\Theta}}^{\lambda_0}_{k}]_{ij}|$$
   \vspace{-4pt}
   \FOR{$\tau_k=0.1\theta_k,0.2\theta_k,\hdots,\theta_k$}
   \vspace{4pt}
   \STATE 1) Hard-threshold $\hat{\vect{\Theta}}^{\lambda_0}_{k}$ element-wise at $\tau_k$: $$\widetilde{\vect{\Theta}}^{(k)}_{\tau_k}=T_{\tau_k}(\hat{\vect{\Theta}}^{\lambda_0}_{k})$$
   \vspace{-10pt}
   \STATE 2) Record the number of edges in $\widetilde{\vect{\Theta}}^{(k)}_{\tau_k}$ as $e_{\tau_k}$
   \vspace{4pt}
   \ENDFOR
   \vspace{4pt}
   \STATE 3) Choose the best threshold level using eBIC criterion:
   \begin{align*}
       \tau_k^{*}=\argmin_{\tau_k\in\{0.1\theta_k,0.2\theta_k,\hdots,\theta_k\}} &-2\ell(\widetilde{\vect{\Theta}}^{(k)}_{\tau_k})\\ &+ e_{\tau_k}\log n + 4e_{\tau_k}\gamma \log m
   \end{align*}
   where $\ell$ is the log-likelihood of the multivariate Gaussian distribution
   \STATE {\bfseries Output:} $\widetilde{\vect{\Theta}}^{(k)}=\widetilde{\vect{\Theta}}^{(k)}_{\tau_k^*}\in\mathbb{R}^{m\times m}$
\end{algorithmic}
\end{algorithm}
}

\textbf{Relation to the StARS Procedure} The Stability Approach to Regularization Selection (StARS) \cite{liu2010stability} is a popular data-driven approach to choosing tuning hyperparameter values for a variety of Gaussian graphical model estimators. Similar to our MPGraph method, StARS repeatedly fits graph estimators to $K$ random subsamples of the observations but includes all the features and then computes a so-called edge instability score by aggregating edge selection events from all subsamples. In contrast, our MPGraph approach employs tiny subsamples of both observations and features, thus necessitating latent variable graphical model estimates but also dramatically reducing the overall computational costs.  Secondly, StARS seeks to find a sparse, stable graph and hence solves the graph selection problem repeatedly at a sequence of tuning hyperparameter values on each subsample. On the other hand, our MPGraph approach performs graph sparsity tuning on each minipatch separately, thus dramatically reducing computational costs.  In the end, our method produces a set of stable edges by simply aggregating over the minipatches with no additional tuning or computation.

\subsection{Theoretical Analysis}
\label{sec:mpgraph-theory}

We have presented a new, fast computational method for learning the structure of Gaussian graphical models.  But since our method is novel and does not solve known Gaussian graphical model selection problems \cite{meinshausen2006high,yuan2007model,cai2011constrained}, one naturally asks: Does MPGraph correctly recover the edge structure of the Gaussian graphical model?  In other words, is our method sparsistent or model selection consistent?  In this section, we establish such theoretical results.

Recall that we employ the thresholded GLasso on each minipatch to obtain a graph estimate.  Recently, \cite{wang2021tglasso} showed that this method is graph selection consistent in the presence of latent variables under certain conditions on the true graph and corresponding covariance matrix as well as the Schur complement of the latent variables.  Then, if we can show that all of these conditions are satisfied for all possible random minipatches, our MPGraph approach will also be graph selection consistent.  This is the approach we take.

First,  denote the true edge set by $E^* = E(\Theta^*) = \{(i,j): \Theta^*_{ij} \neq 0 \:\text{and}\:i\neq j\}$ and $s = \max_{k} s_k$ where  $s_k = |E \left(\Theta_{F_k}^{*}\right)|$ refers to the total number of non-zero edges for the $k$th minipatch $F_k$. Also let $\varphi_{\max}()$ and $\varphi_{\min}()$ denote the maximum and minimum eigenvalue respectively. Denote the elementwise $\ell_{\infty}$-norm of a matrix by $\|\mathbf{A}\|_{\infty}=\max_{i\neq j}|A_{ij}|$.
 Consider the following assumptions:
\begin{assumption}[restate = assumptionMPGone, name = ]
\label{assumptionMPG1}
$X_{i} \text { be i.i.d. } \mathcal{N}\left(\mathbf{0}, \Sigma^* \right)$.
\end{assumption}
\vspace{0.5em}
\begin{assumption}[restate = assumptionMPGtwo, name = ]
\label{assumptionMPG2}
 $\varphi_{\min }\left( \Sigma^* \right) \geq \underline{\kappa}>0$.
\end{assumption}

\vspace{0.5em}
\begin{assumption}[restate = assumptionMPGthree, name = ]
\label{assumptionMPG3}
$\varphi_{\max }\left(\Sigma^*\right) \leq \overline{\kappa}$.
\end{assumption}

\vspace{0.5em}
\begin{assumption}[restate = assumptionMPGfour, name = ]
\label{assumptionMPG4}
Minimum edge strength:
\begin{equation*}
\theta_{\min }:=\min _{(i, j) \in E\left(\bTheta^{*}\right)}\left|\bTheta_{i j}^{*}\right| > c_1 \sqrt{\frac{s \log m}{n}}.
\end{equation*}

\end{assumption}

\vspace{0.5em}
\begin{assumption}[restate = assumptionMPGfiveraw, name = ]
\label{assumptionMPG6}
Maximum effect of unsampled nodes:
\begin{align*}
&\max \limits_{F_k} \|  \left( \bTheta_{F_k}^{*}  -  \bTheta_{F_k F_k^c}^{*} \left(\bTheta_{F_k^c}^{*}\right)^{-1} \bTheta_{F_k^c F_k}^{*} \right)^{-1} - \left(\bTheta_{F_k}^{*}\right)^{-1}   \|_{\infty}  \\ &=
\mathcal{O}\left(\sqrt{\frac{\log m}{n}}\right).
\end{align*}
\end{assumption}

Further, let $R_{ij}$ denote the number of times any pair of nodes $i,j$ are sampled together into minipatches.  Define constant $\nu$ to satisfy $m^{\nu} = b_1 \exp(-b_2 n \lambda^2)$ where $b_1$ and $b_2$ depend on $\overline{\kappa}$ 
(defined in the Supplement); also, define $b_0 = \min \left\{ \frac{(1-m^{\nu} - \pi_{\text{thr}})^2 }{2},   \frac{( \pi_{\text{thr}}-m^{\nu} )^2 }{2} \right\}$; finally, define the constant $L$ that satisfies $|S_{ij}^{(k)}  - (1-m^{\nu}) | \leq L$ for all pairs of $(i,j) \in E^*$ that are sampled on the $k$th minipatch and $|S_{ij}^{(k)}  - m^{\nu} | \leq L$ for all pairs of $(i,j) \in {E^*}^c$ that are sampled on the $k$th minipatch.
Given these preliminaries, we are able to state our main theoretical result:

\begin{theorem}
\label{MPGtheorem}
Let Assumptions~\ref{assumptionMPG1}-\ref{assumptionMPG6}  be satisfied and let $n$ grow proportionally with $N$. Then, the minipatch graph selection estimator, MPGraph, with $\lambda \asymp \sqrt{\frac{\log m}{n}}$ and $\tau \asymp \sqrt{\frac{s \log m}{n}}$, is graph selection consistent with high probability:
\end{theorem}
\begin{align*}
    \mathbb{P}\left(  \hat{E}^{\text{stable}} =  E^*\right)
    &\geq 1 - 2m^2 \exp \left\{ - b_0 \cdot \frac{\min_{i,j} R_{ij}}{m^{\nu} + L  / 3 } \right\}  \\
    & \to 1 , \hspace{6mm} \text{ as } N \to \infty,  \text{ or as }
    K \to \infty.
\end{align*}

Theorem~\ref{MPGtheorem} establishes finite-sample graph selection consistency of our MPGraph method with high probability. We provide a detailed proof in the Supplementary Material, but pause to briefly discuss this result. First, notice that the probability of correct graph selection tends to one as $N \to \infty$ or $K \to \infty$.  To see this, notice that $m^{\nu} \in (0,1)$ and $b_0$ is bounded. Then, as the number of minipatches, $K$, increases, so does $min_{i,j} \ R_{ij}$, thus establishing the result when $K \to \infty$.  On the other hand for fixed $K$, notice that $m^{\nu} \to 0$ as $N \to \infty$ and hence $L \to 0$ also as $N \to \infty$; this also establishes our result.

Assumptions~\ref{assumptionMPG4} and~\ref{assumptionMPG6} are the key assumptions ensuring the overall graph selection consistency of our MPGraph approach. Assumption~\ref{assumptionMPG6} suggests that the effect of unsampled (latent) variables should not exceed the order of $\mathcal{O}(\sqrt{\log m / n})$. This assumption ensures that the false edges induced by unsampled (latent) variables are small in magnitude, thus allowing the thresholding procedure to successfully remove them. In practice, we have some control to ensure Assumption~\ref{assumptionMPG6} holds by making sure the number of sampled nodes $m=|F_k|$ in the minipatches is not too small relative to $n$.
We investigate the effect of $m$ further and when Assumption~\ref{assumptionMPG6} holds in Sec.~\ref{sec:mpgraph-valid}).
Interestingly, Assumption~\ref{assumptionMPG6} also suggests that there is a trade-off between the statistical accuracy and computational efficiency for our method. When $n$ and $m$ are sufficiently large, our approach will satisfy this assumption and enjoy better statistical guarantees, but clearly this will result in a larger computational burden as well.
Intuitively, Assumption~\ref{assumptionMPG4} requires the minimum edge strength of the true graph to be sufficiently large in magnitude such that the thresholding procedure will eliminate false positive edges while not removing true edges on a minipatch.

Notice that our MPGraph algorithm does not require an irrepresentability condition for graph selection consistency; this differs from the graphical Lasso and similar algorithms which require this stringent assumption.  Instead, our MPGraph algorithm assumes that the eigenvalues of the covariance matrix are bounded (Assumption~\ref{assumptionMPG2} and ~\ref{assumptionMPG3}), a significantly weaker assumption.
Additionally, it is worth noting that we do not require low-rankness for the unsampled (latent) component as in \cite{chandrasekaran2012} because we employ the thresholded GLasso \cite{wang2021tglasso} approach on each minipatch.

\begin{figure*}[!ht]
\centering
\includegraphics[width=0.9\linewidth]{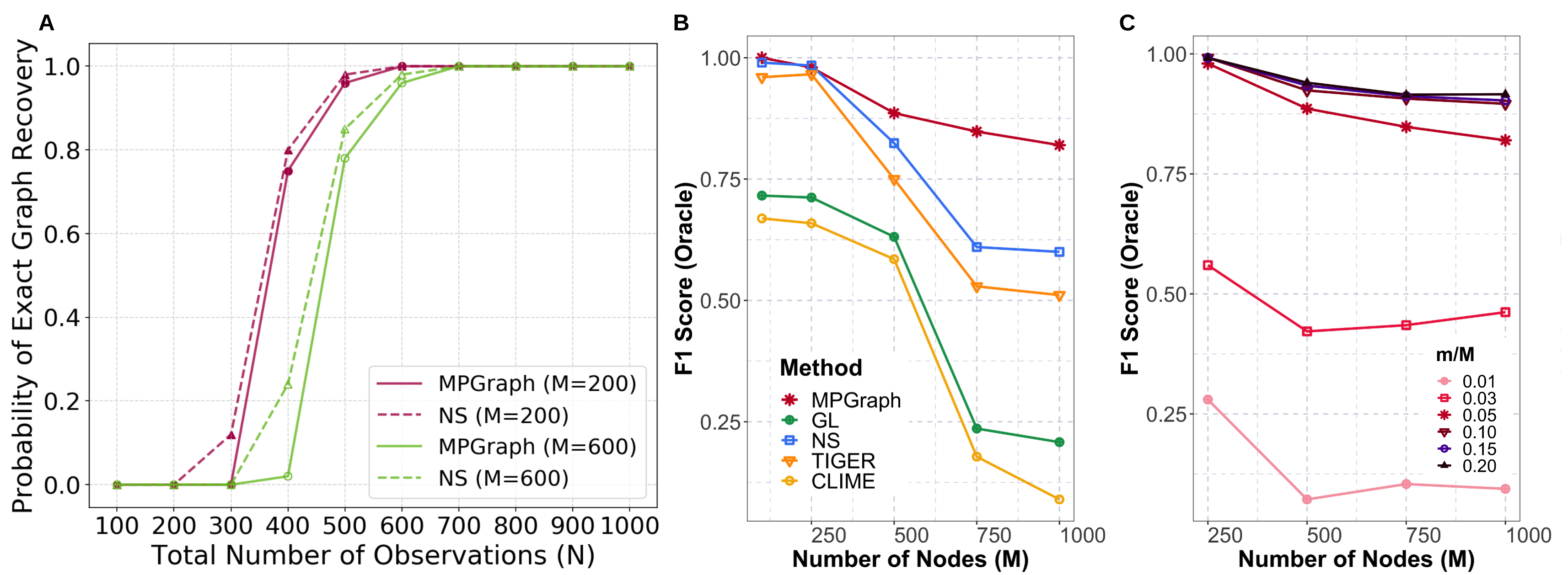}
\caption{Validation of Theoretical Results. (A) Probability of exact edge-set recovery versus total sample size $N$ for chain graph simulations with varying number of nodes $M$. Each point represents the average over $100$ trials. (B) Edge selection accuracy (F1 Score) comparisons for small-world graph simulations with fixed $N=500$ and varying $M$. Note that oracle parameter tuning is used for all methods. (C) Edge selection accuracy of MPGraph using various minipatch sizes (i.e. $m/M$) for the same simulations in (B).}
\label{fig:sample_complexity}
\end{figure*}

Finally, Theorem~\ref{MPGtheorem} suggests that the sample complexity required for our MPGraph estimator is $n = \mathcal O(s \log m)$ for each minipatch. We expect that the overall sample complexity is of the same order as neighborhood selection and verify this empirically in Figure~\ref{fig:sample_complexity}A.

Overall, our theoretical result provides the reassurance that one can correctly estimate the graph structure with high probability using ensembles of minipatches.  One can also argue that our assumptions are significantly weaker than those required by the graphical Lasso for graph selection consistency.  Hence, we expect that our approach will not only perform well computationally, but also offer major statistical advantages as well.  We seek to empirically validate our theory and investigate this claim in the next section.

\subsection{Practical Considerations}\label{sec:param}

Our MPGraph is a general meta-algorithm with three tuning hyperparameters: minipatch size ($n$ and $m$) and threshold $\pi_{\text{thr}}$. In practice, our method has quite robust performance for sensible choices of these hyperparameters. We have additional empirical experiments in the Supplementary Material investigating how edge selection accuracy and computational time of MPGraph vary for various minipatch sizes. As a general rule of thumb, we found taking $m$ to be $5\%\sim 10\%$ of $M$ and then picking $n$ relative to $m$ so that it surpasses the sample complexity of the base graph estimator used in the meta-algorithm strikes a good balance between statistical accuracy and computational time. In addition, we found fixing the threshold $\pi_{\text{thr}}$ to $0.5$ and the total number of iterations $K=1000$ works well for many problems in practice. Note that the internal latent variable graph selection estimator also has tuning parameters, in our case the threshold level $\tau_k$, but we strongly suggest tuning these separately for each minipatch (as we outline in Algorithm~\ref{alg:thresholdedglasso}) and not all together across minipatches.

\section{Empirical Studies}
\label{sec:mpgraph-empirical}

In this section, we follow standard computational benchmarks \cite{liu2010stability,hsieh2013bigquic,muller2016generalized} to compare our proposed MPGraph method with a wide range of existing Gaussian graphical model selection strategies in terms of both edge selection accuracy and computational time. In particular, we consider the following widely-used groundtruth graph structures for generating synthetic benchmark data sets:

\begin{itemize}
    \item {\bf Scenario 1: Chain graph:} The groundtruth precision matrix $\vect{\Theta}\in\mathbb{S}^{M\times M}_{+}$ is set to be $\Theta_{i,i-1}=\Theta_{i-1,i}=0.6$ and $\Theta_{i,i}=1.25$ for all $i=1,\hdots,M$.
    \item {\bf Scenario 2: Erd\H{o}s-Renyi graph:} First generate an adjacency matrix $\mathbf{A}\in\{0,1\}^{M\times M}$ whose edge patterns follow a Erd\H{o}s-Renyi graph with edge probability $p$ such that the total number of undirected edges is $(M-1)$. Then the groundtruth precision matrix $\vect{\Theta}$ is set to be $\Theta_{i,j}=\Theta_{j,i}=\rho_{ij}$ if $A_{i,j}=1$ and $\Theta_{i,j}=\Theta_{j,i}=0$ otherwise, where $\rho_{ij}\sim \text{uniform}([-0.6,-0.3]\cup [0.3,0.6])$. Lastly, set $\Theta_{i,i}=1$ for all $i=1,\hdots,M$.
    \item {\bf Scenario 3: Small-World graph:} First generate an adjacency matrix $\mathbf{A}\in\{0,1\}^{M\times M}$ whose edge patterns follow a Watts-Strogatz small-world graph with $2$ nearest neighbors and edge rewiring probability of $0.5$. Then the groundtruth precision matrix $\vect{\Theta}$ is generated in the same way as in the Erd\H{o}s-Renyi graph scenario.
\end{itemize}

For a given groundtruth precision matrix $\vect{\Theta}$, we generate the data matrix $\mathbf{X}\in\mathbb{R}^{N\times M}$ whose rows are independently drawn from a $M$-variate Gaussian distribution $\mathcal{N}(\mathbf{0}, \vect{\Theta}^{-1})$. The groundtruth edge set is $E=\{1\leq i < j \leq M: \Theta_{i,j}\neq 0\}$. We evaluate the edge selection accuracy of each method in terms of the F1 Score, which takes values between $0$ and $1$ with $1$ signifying perfect match between the estimated edge set and the groundtruth edge set $E$. We consider a range of low-dimensional and high-dimensional experimental scenarios with various $N$ and $M$ sizes in Sec.~\ref{sec:mpgraph-valid}-\ref{sec:mpgraph-empirical-large}. All comparisons were conducted on a VM with $12$ vCPUs (Intel Cascade Lake) with $240$ GB of memory.

\subsection{Validation of Theoretical Results}
\label{sec:mpgraph-valid}

Before conducting thorough comparative studies in Sec.~\ref{sec:mpgraph-empirical-small}-\ref{sec:mpgraph-empirical-large}, we want to first empirically illustrate and verify the theoretical properties of MPGraph. Following similar experiments in \cite{ravikumar2011high}, we show the probability of exact edge-set recovery versus sample size $N$ for chain graph simulations with varying number of nodes $M$ in Figure~\ref{fig:sample_complexity}A. We see that the recovery probability of MPGraph follows that of neighhorhood selection (NS) closely, and reaches 1 at the same $N$ as NS. This supports our theoretical analysis as it suggests the overall sample complexity of MPGraph is of the same order as NS for consistent graph selection. In addition, we compare the edge selection accuracy of MPGraph to standard GGM selection procedures using oracle tuning approaches (i.e. assume the total number of true edges $|E|$ is known) in Figure~\ref{fig:sample_complexity}B. We see that even though accuracy of all methods degrades as the graph dimension $M$ increases, MPGraph generally achieves better F1 Score even in high-dimensional settings where exact edge-set recovery might not be attainable.

In Figure~\ref{fig:sample_complexity}C, we verify that the number of nodes in a minipatch $m$ is the determining factor for how well MPGraph recovers the true graph because it dictates whether Assumptions~\ref{assumptionMPG4} and~\ref{assumptionMPG6} would be largely satisfied in practice. For the challenging small-world graph situations in Figure~\ref{fig:sample_complexity}C, Assumption~\ref{assumptionMPG6} is likely not satisfied when $m$ is too small, hence performance of MPGraph degrades precipitously. But with a sufficiently large $m$ (e.g. at least $5\%$ of $M$), we can achieve better performance. More results are in the Supplementary Material. Thus, we recommend choosing $m$ to be $5\%\sim 10\%$ of $M$ and then picking $n$ relative to $m$ so that it surpasses the sample complexity of the base graph estimator used on the minipatches, which would allow ample room for Assumption~\ref{assumptionMPG4} and~\ref{assumptionMPG6} to be satisfied for reasonable graphs in practice.

\begin{figure*}[!ht]
\centering
\includegraphics[width=0.8\linewidth]{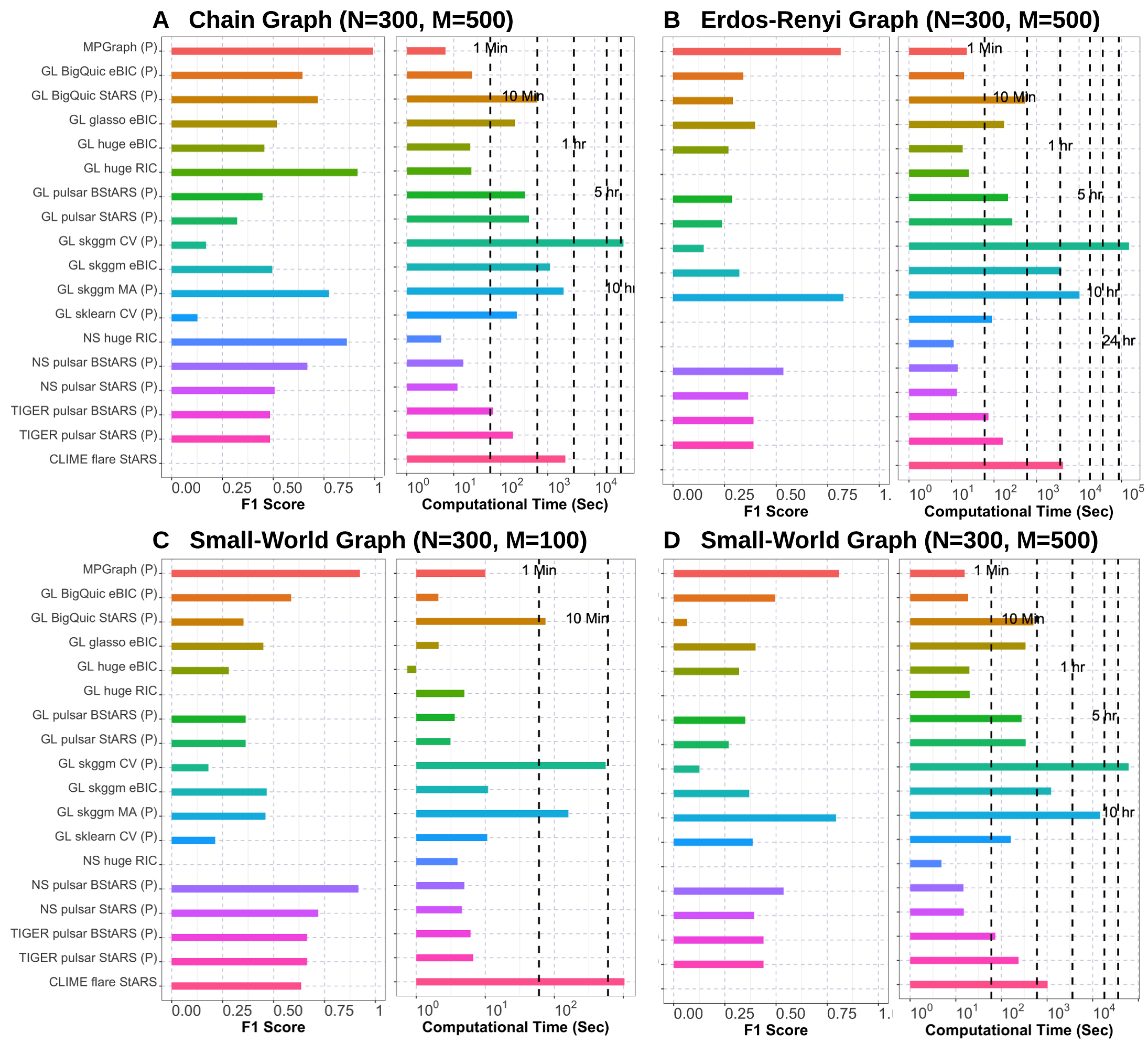}
\caption{Edge Selection Accuracy (F1 Score) and Computational Time from Simulation Scenarios 1-3 for a Variety of Dimensions. Parallelism is enabled for all methods whose software packages include this functionality, as indicated by the (P) after the method name. Our MPGraph method achieves the best edge selection accuracy across all four data sets while being one of the computationally fastest methods.}
\label{fig:smallscale}
\end{figure*}

\subsection{Comparative Empirical Studies}
\label{sec:mpgraph-empirical-small}

In this section, our goal is to empirically compare our proposed MPGraph method to a wide range of existing computational approaches and state-of-the-art solvers implemented in widely-used software packages for Gaussian graphical model selection.  The approaches we consider include the sparse inverse covariance estimation (GL), neighborhood selection (NS), CLIME, and TIGER. In particular, for the GL approach, we consider the QUIC algorithm \cite{hsieh2014quic} (\texttt{skggm} \cite{laska_narayan_2017_830033}), the BigQUIC algorithm (\texttt{BigQuic}) \cite{hsieh2013bigquic}, and various coordinate-descent-based algorithms (\texttt{glasso} \cite{friedman2008sparse}, \texttt{huge} \cite{zhao2012huge}, \texttt{pulsar} \cite{muller2016generalized}, \texttt{sklearn} \cite{scikit-learn}). In addition, we use both \texttt{huge} and \texttt{pulsar} for NS, \texttt{pulsar} for TIGER, and \texttt{flare} \cite{flare} for CLIME. Our MPGraph algorithm is implemented in \texttt{Python}.

\columnalignment{tag1}{r}
\begin{table*}[!hb]
  \caption{Experimental Results for Large-Scale Gaussian Graphical Model Selection Problems. The number of selected edges, true positive rate (TPR), precision, F1 Score, and computational time in seconds are reported for each method. Note that parallelism is enabled for all methods whose software implementations include this functionality, as indicated by the (P) after the method name. The best F1 Score and runtime for each scenario are bold-faced. Our MPGraph method achieves the best edge selection accuracy (F1 Score) across all scenarios, and it is the computationally the fastest, especially for huge graph selection problems with tens of thousands of nodes.}
  \label{tab:largescale}
  \centering
  \begin{tabular}{clccccc}
    \toprule
    \multicolumn{1}{c}{} & \multicolumn{1}{c}{Method} &\multicolumn{1}{c}{ \# Selected Edges} & \multicolumn{1}{c}{TPR} & \multicolumn{1}{c}{Precision} & \multicolumn{1}{c}{F1 Score} & \multicolumn{1}{c}{Time (S)} \\
    \midrule
    \multirow{6}{*}{ \vtop{\hbox{\strut Small-World}\hbox{\strut $N=500$}\hbox{\strut $M=1000$ \space\space\space}} } & MPGraph (P) & 1387 & 0.882 & 0.636 & \textbf{0.739} & \textbf{14.842} \\
    & GL huge eBIC & 4872 & 0.996 & 0.204 & 0.339 & 135.399 \\
    & NS pulsar BStARS (P) & 2767 & 0.999 & 0.361 & 0.530 & 41.388 \\
    & GL BigQuic eBIC (P) & 1697 & 0.822 & 0.484 & 0.610 & 76.097 \\
    & GL glasso eBIC & 3910 & 0.956 & 0.245 & 0.389 & 3217.082 \\
    & GL skggm MA (P) & 1009 & 0.665 & 0.659 & 0.662 & 3794.389 \\
    \bottomrule
    \toprule
    \multirow{6}{*}{ \vtop{\hbox{\strut Erdos-Renyi}\hbox{\strut $N=500$}\hbox{\strut $M=1000$ \space\space\space}} } &  MPGraph (P) & 1112 & 0.822 & 0.730 & \textbf{0.773} & \textbf{20.403} \\
    & GL huge eBIC & 7079 & 0.994 & 0.139 & 0.243 & 120.357 \\
    & NS pulsar BStARS (P) & 3022 & 0.999 & 0.327 & 0.492 & 41.488 \\
    & GL BigQuic eBIC (P) & 2186 & 0.744 & 0.336 & 0.463 & 73.211 \\
    & GL glasso eBIC & 4592 & 0.921 & 0.198 & 0.326 & 1716.447 \\
    & GL skggm MA (P) & 763 & 0.484 & 0.626 & 0.546 & 4010.957 \\
    \bottomrule
    \toprule
    \multirow{6}{*}{ \vtop{\hbox{\strut Chain Graph}\hbox{\strut $N=500$}\hbox{\strut $M=1000$ \space\space\space}} } & MPGraph (P) & 1027 & 0.991 & 0.964 & \textbf{0.977} & \textbf{9.546} \\
    & GL huge eBIC & 3719 & 1.000 & 0.269 & 0.423 & 154.775 \\
    & NS pulsar BStARS (P) & 1785 & 1.000 & 0.560 & 0.718 & 42.757 \\
    & GL BigQuic eBIC (P) & 2074 & 1.000 & 0.482 & 0.650 & 106.036 \\
    & GL glasso eBIC & 2929 & 1.000 & 0.341 & 0.509 & 1878.841 \\
    & GL skggm MA (P) & 1394 & 0.998 & 0.715 & 0.833 & 1437.701 \\
    \bottomrule
    \toprule
    \multirow{6}{*}{ \vtop{\hbox{\strut Chain Graph}\hbox{\strut $N=2500$}\hbox{\strut $M=5000$ \space}} } & MPGraph (P) & 4594 & 0.919 & 1.000 & \textbf{0.957} & \textbf{85.806} \\
    & GL huge eBIC & 14686 & 1.000 & 0.340 & 0.508 & 28731.445 \\
    & NS pulsar BStARS (P) & 17716 & 1.000 & 0.282 & 0.440 & 3062.926 \\
    & GL BigQuic eBIC (P) & 11490 & 1.000 & 0.435 & 0.606 & 5815.430 \\
    & GL glasso eBIC & - & - & - & - & $>8$ hours \\
    & GL skggm MA (P) & 7597 & 1.000 & 0.658 & 0.794 & 17011.503 \\
    \bottomrule
    \toprule
    \multirow{6}{*}{ \vtop{\hbox{\strut Chain Graph}\hbox{\strut $N=5000$}\hbox{\strut $M=10000$}} } & MPGraph (P) & 9162 & 0.916 & 1.000 & \textbf{0.956} & \textbf{351.143} \\
    & GL huge eBIC & - & - & - & - & $>8$ hours \\
    & NS pulsar BStARS (P) & 61372 & 1.000 & 0.163 & 0.280 & 26737.656 \\
    & GL BigQuic eBIC (P) & 24321 & 1.000 & 0.411 & 0.583 & 27157.235 \\
    & GL glasso eBIC & - & - & - & - & $>8$ hours \\
    & GL skggm MA (P) & - & - & - & - & $>8$ hours \\
    \bottomrule
    \toprule
    \multirow{6}{*}{ \vtop{\hbox{\strut Chain Graph}\hbox{\strut $N=10000$}\hbox{\strut $M=20000$}} } & MPGraph (P) & 18230 & 0.912 & 1.000 & \textbf{0.954} & \textbf{1680.658} \\
    & GL huge eBIC & - & - & - & - & $>8$ hours \\
    & NS pulsar BStARS (P) & - & - & - & - & $>8$ hours \\
    & GL BigQuic eBIC (P) & - & - & - & - & $>8$ hours \\
    & GL glasso eBIC & - & - & - & - & $>8$ hours \\
    & GL skggm MA (P) & - & - & - & - & $>8$ hours \\
    \bottomrule
  \end{tabular}
\end{table*}

In order to mimic realistic use cases of Gaussian graphical model selection in practice, we do not assume the oracle optimal tuning hyperparameter is known. Instead, widely-used tuning hyperparameter selection schemes from the aforementioned software packages are used to choose the optimal hyperparameter values for each method in a data-driven manner. Specifically, a geometric sequence of candidate hyperparameter values $\Lambda$ of length $100$ in the range of $(\delta\lambda_{\text{max}}, \lambda_{\text{max}})$ are used for all software packages, where the ratio $\delta\in(0,1)$ is set to the value recommended by the respective software package. Additionally, we use the suggested minipatch size (i.e. $m/M=10\%$ and $m/n=0.8$) and default settings (i.e. $\pi_{\text{thr}}=0.5$ and $K=1000$) for MPGraph for comparisons in this section. Last but not least, parallelism is enabled for all methods whose software implementations include this functionality.

In Figure~\ref{fig:smallscale}, we display the performance of all methods on the three graph types for a variety of dimensions. At a quick glance, MPGraph outperforms most of the competing methods in terms of edge selection accuracy across all four data sets. While some state-of-the-art methods such as GL BigQuic StARS, NS pulsar B-StARS, and GL huge RIC attain the same or similar edge selection accuracy to MPGraph in scenarios with simpler graph structure (Figure~\ref{fig:smallscale}A) or low-dimensional settings (Figure~\ref{fig:smallscale}C), their accuracy deteriorates considerably when the complexity of the underlying graph structure and/or the dimension of the problem increase (Figure~\ref{fig:smallscale}B,~\ref{fig:smallscale}D). On the other hand, the edge selection accuracy of MPGraph does not degrade as much when the underlying graphical model selection problem becomes more challenging. In the meantime, MPGraph is one of the computationally fastest methods, with computational time on the order of seconds to a minute as opposed to hours required by many competitors with similar accuracy (e.g. GL skggm MA), especially in the more challenging scenarios (Figure~\ref{fig:smallscale}B,~\ref{fig:smallscale}D). Even though MPGraph is designed for tackling huge graph selection problems, it is still a solid choice for smaller-scale data sets in practice.

\subsection{Large-Scale Comparative Studies}
\label{sec:mpgraph-empirical-large}

In this section, we consider the problem of estimating the structure of large-scale Gaussian graphical models with the number of nodes ranging from thousands to tens-of-thousands, similar to the experiments in \cite{hsieh2014quic}. Problems of such scale are often encountered in neuroscience and genomics, among many others. To save computational resources, we only include the most accurate and/or the fastest competitors from Sec~\ref{sec:mpgraph-empirical-small} for these large-scale comparisons. In addition, following similar practices in \cite{hsieh2014quic}, we stop any method whose runtime exceeds 8 hours. The same data-driven hyperparameter tuning schemes as in Sec.~\ref{sec:mpgraph-empirical-small} are used for these methods. The default minipatch size (i.e. $m/M=5\%$ and $m/n=0.8$) and settings (i.e. $\pi_{\text{thr}}=0.5$ and $K=1000$) are used for MPGraph in this section. Additionally, parallelism is enabled for all methods whose software implementations include this functionality.

The experimental results are summarized in Table~\ref{tab:largescale}. We see that our MPGraph method consistently obtains the highest edge selection accuracy across all scenarios, often leading the next best performing method by a large margin. Taking a closer look, we see that one big advantage of MPGraph is that it consistently selects considerably fewer false edges (as indicated by its higher precision) than the competing methods while still identifying most of the true edges (as indicated by its decent TPR). This can have a big impact in many real-world applications. For instance, having fewer false edges in the estimated graph in genomics research means huge savings in time and resources that would otherwise be spent on investigating false relationships between genes. At the same time, our MPGraph is computationally the fastest across the board, and it is extensively faster than the state-of-the-art competitors for huge graph selection problems, taking less than 30 minutes for estimating a graph with $M=20000$ nodes.

Overall, we empirically demonstrate that our proposed MPGraph method could be a win both in terms of edge selection accuracy and computation in practice, especially for huge graph selection problems. Intuitively speaking, the empirical effectiveness of our approach can be attributed to the following factors: i) splitting up huge (high-dimensional) graph selection problems into smaller (lower-dimensional) ones (minipatches) generally leads to faster and more accurate graph estimation by each individual base estimator; ii) fitting these base graph estimators over minipatches in parallel further yields major computational savings; iii) the final ensemble of edge selection frequency over all minipatches reduces variance in a similar manner to that of stability-based approaches and provides an effective guard against false edges, hence leading to more accurate graph structure estimations.

\section{Case Study: Neuroscience Data}

We now study the performance of our MPGraph method on a real-world large-scale calcium imaging data set from a neuroscientific study of the mouse primary visual cortex \cite{STRINGER20212767}. Specifically, the data set recorded activities of $M=14515$ neurons (nodes) over $N=4279$ time points (observations) during which square static gratings with various orientations were shown to the mouse (see \cite{STRINGER20212767} for detailed data acquisition procedures). The data had been deconvolved and temporally decorrelated such that it is suitable for various downstream statistical analysis. One important area of research in neuroscience is the analysis of statistical relationships, or functional connectivity, between individual neurons in the brain. As a powerful approach for learning conditional dependence relationships, graphical models are a popular choice for carrying out such analysis in calcium imaging data \cite{yatsenko2015improved, chang2019graphical, vinci2019graph}. For this case study, we apply our MPGraph method to this challenging large-scale, high-dimensional data set to estimate the functional neuron connectivity network. We also include the BigQUIC algorithm and the neighborhood selection procedure for performance comparisons, which are the state-of-the-art computational approaches to Gaussian graphical model selection.

Because the groundtruth edge set (or the groundtruth structure of the functional connectivity network) is unknown in such real-world data set, we evaluate the edge selection performance with respect to neural orientation tunings. Such evaluation criterion has been widely used in the graphical model literature \cite{chang2021extreme,wang2021tglasso}. Specifically, neuroscientific studies have suggested that neurons that are tuned to the same stimuli are more likely to be connected functionally \cite{sakai1994}. Therefore, the percentage of edges that are between neuron pairs with the same orientation tuning is used to evaluate the functional connectivity graph estimated by each method. Following \cite{chang2021extreme,wang2021tglasso}, a higher percentage suggests that the final estimated graph is likely to be more neuroscientifically plausible.

To mimic realistic use cases in practice, the same types of data-driven hyperparameter selection procedures as in Sec.~\ref{sec:mpgraph-empirical-large} are used for BigQUIC and neighborhood selection, as the optimal tuning hyperparameter values are generally unknown a priori in these types of real-world data applications. On the other hand, MPGraph is run with the default settings (i.e. $m/M=5\%$ and $m/n=0.8$, $\pi_{\text{thr}}=0.5$ and $K=1000$).

We display the quantitative experimental results in Table~\ref{mpgraph-real-data-1}, \ref{mpgraph-real-data-2}, and \ref{mpgraph-real-data-3}. Specifically, we show the number of selected edges, the percentage of edges that are between neuron pairs with the same tuning, and computational time in seconds for each method. Note that parallelism is enabled for all methods whose software implementations include this functionality, as indicated by the (P) after the method name. The best percentage and runtime in each scenario are bold-faced.

From Table~\ref{mpgraph-real-data-1}, we can see that $70.6\%$ of the edges in the functional connectivity graph estimated by MPGraph are between neurons with the same orientation tuning, which is higher than that of the graphs estimated by the competing methods. This suggests that the functional connectivity estimate from MPGraph is perhaps more scientfically appropriate compared to these from the other methods. At the same time, MPGraph is computationally the fastest, taking less than 15 minutes as opposed to the 11.6 hours required by the BigQUIC algorithm with data-driven hyperparameter tuning.

\columnalignment{tag1}{r}
\begin{table}[H]
\caption{Experimental Results on the Calcium Imaging Data with Data-Driven Hyperparameter Tuning.}
\label{mpgraph-real-data-1}
\begin{center}
  \scalebox{0.9}{\parbox{1\linewidth}{%
  \begin{tabular}{lccc}
    \toprule
    \multicolumn{1}{c}{Method} &\multicolumn{1}{c}{ \# Selected Edges} & \multicolumn{1}{c}{Percentage (\%)} &  \multicolumn{1}{c}{Time (S)} \\
    \midrule
    MPGraph (P) & 1631 & \bfseries 70.6 & \bfseries 662.894 \\
    GL BigQuic eBIC (P) & 37351 & 63.5 & 41921.747 \\
    NS pulsar BStARS (P) & 44900 & 54.6 & 49169.265 \\
    \bottomrule
  \end{tabular}
  }}
\par
\end{center}
\end{table}

One advantage of our MPGraph approach is that it has integrated stability-based hyperparameter tuning, which does not require separate tuning procedures as in the case of most existing Gaussian graphical model selection techniques such as BigQUIC and NS. However, for the sake of argument, let us assume that the optimal tuning hyperparameter values determined by the eBIC and the BStARS procedures are already known for BigQUIC and NS, respectively. We subsequently record the computational time of these competitors fit at a single hyperparameter value, as shown in Table~\ref{mpgraph-real-data-2}. We see that MPGraph is still noticeably faster even when the computationally burdensome tuning procedures for the competing methods are taken out of the equation.

\columnalignment{tag1}{r}
\begin{table}[H]
\caption{Experimental Results on the Calcium Imaging Data with a Single Optimally Tuned Hyperparameter Value for Competing Methods. }
\label{mpgraph-real-data-2}
\begin{center}
  \scalebox{0.9}{\parbox{1\linewidth}{%
  \begin{tabular}{lccc}
    \toprule
    \multicolumn{1}{c}{Method} &\multicolumn{1}{c}{ \# Selected Edges} & \multicolumn{1}{c}{Percentage (\%)} &  \multicolumn{1}{c}{Time (S)} \\
    \midrule
    MPGraph (P) & 1631 & \bfseries 70.6 & \bfseries 662.894 \\
    GL BigQuic $\lambda_{\text{eBIC}}^{*}$ (P) & 37351 & 63.5 & 986.445 \\
    NS pulsar $\lambda_{\text{BStARS}}^{*}$ (P) & 44900 & 54.6 & 1090.994 \\
    \bottomrule
  \end{tabular}
  }}
\par
\end{center}
\end{table}

Some might argue that it would be more fair to compare the percentage of edges that are between neuron pairs with the same tuning if the estimated graphs have similar number of edges (or sparsity level). To this end, we manually find the single tuning hyperparameter value that results in similar number of estimated edges to MPGraph for the competing methods. The experimental results are reported in Table~\ref{mpgraph-real-data-3}. As we can see, even when the sparsity levels of the estimated graphs are similar, MPGraph still outperforms the competitors in terms of both edge selection performance and computational speed. This provides us with further evidence that the functional connectivity network estimate by MPGraph is more neuroscientifically plausible, lending support to the practical utility of our proposed approach.

\columnalignment{tag1}{r}
\begin{table}[H]
\caption{Experimental Results on the Calcium Imaging Data with a Single Hyperparameter Value that Gives Similar Sparsity Level in the Estimated Graph. }
\label{mpgraph-real-data-3}
\begin{center}
  \scalebox{0.9}{\parbox{1\linewidth}{%
  \begin{tabular}{lccc}
    \toprule
    \multicolumn{1}{c}{Method} &\multicolumn{1}{c}{ \# Selected Edges} & \multicolumn{1}{c}{Percentage (\%)} &  \multicolumn{1}{c}{Time (S)} \\
    \midrule
    MPGraph (P) & 1631 & \bfseries 70.6 & \bfseries 662.894 \\
    GL BigQuic $\lambda_{\text{GL}}^{*}$ (P) & 1688 & 66.0 & 742.441 \\
    NS pulsar $\lambda_{\text{NS}}^{*}$ (P) & 1836 & 59.7 & 1085.398 \\
    \bottomrule
  \end{tabular}
  }}
\par
\end{center}
\end{table}

\section{Discussion}

In this work, we have developed a statistically accurate and fast computational approach to learn the structure of Gaussian graphical models named MPGraph that leverages both minipatch learning and thresholded graph estimators. We empirically demonstrate that our approach is not only more accurate but also extensively faster for huge graph selection problems. Additionally, we theoretically analyze MPGraph showing that it attains graph selection consistency under weaker conditions than many existing graph estimators, lending theoretical support to the strong empirical performance of our method. We also provide a high-quality implementation of our proposed method in the open-source \texttt{Python} package \texttt{minipatch-learning} to enable the application of MPGraph to more data scenarios in practice.

Future methodological work might involve developing adaptive sampling schemes, following ideas from \cite{yao2021feature}. In future theoretical work, one could perhaps extend our graph selection consistency results to provide bounds on the number of falsely selected edges or perhaps conduct inference and control the false discovery rate of the selected edges. Further, another interesting direction for future research is to study whether our approach can be extended to other large-scale graphical models such as the Ising model and the exponential family graphical model.

\section*{Acknowledgments}
\noindent The authors acknowledge support from NSF DMS-2210837, NSF NeuroNex-1707400, and NIH 1R01GM140468.

\newpage

\bibliographystyle{IEEEtran}
\bibliography{./mybib.bib}

\newpage

\onecolumn
\begin{center}
{\bf \LARGE Fast and Accurate Graph Learning for Huge Data via Minipatch Ensembles: Supplementary Materials}
\bigskip

{\large Tianyi Yao, Minjie Wang, and Genevera I. Allen}
\end{center}

\renewcommand\thesection{\Alph{section}}
\renewcommand\thesubsection{\thesection.\arabic{subsection}}


\setcounter{section}{0}
\setcounter{assumption}{0}
\setcounter{theorem}{0}

\setcounter{figure}{3}

\section{Theoretical Analysis}
\label{sec:mpgraph-proof}

In this section, we prove the theoretical properties of our MPGraph approach established in Sec. II-B of the main paper. In particular, we show the graphical model selection consistency of our method.

\vspace{0.5em}
We first show that our estimate is graph selection consistent on each minipatch; hence our estimate is consistent aggregating over all minipatches. For each minipatch, we treat the sampled nodes as observed nodes, and unsampled nodes as unobserved latent variables, leading to the problem of graph selection in the presence of latent variables.

\vspace{0.5em}
Denote $F_k$ as the minipatch set for features.
The inverse covariance matrix $\bTheta = \Sigma^{-1}$ can be written as
$\bTheta=\left(\begin{array}{cc}
\bTheta_{F_k} & \bTheta_{F_k F_k^c} \\
\bTheta_{F_k^c F_k} & \bTheta_{F_k^c}
\end{array}\right).$
The marginal concentration matrix $(\Sigma^*_{F_k})^{-1}$ associated with the sampled features $X_{F_k}$, is given by the Schur complement:
$$
\left(\Sigma_{F_k}^{*}\right)^{-1}=\bTheta_{F_k}^{*}-\bTheta_{F_k F_k^c}^{*}\left(\bTheta_{F_k^c}^{*}\right)^{-1} \bTheta_{F_k^c F_k}^{*},
$$

\vspace{0.5em}
We restate the assumptions we have made in Sec. II-B of the main paper. These assumptions guarantee that the assumptions in \cite{wang2021tglasso} are satisfied for each minipatch. First,  denote the true edge set by $E^* = E(\Theta^*) = \{(i,j): \Theta^*_{ij} \neq 0 \:\text{and}\:i\neq j\}$ and $s = \max_{k} s_k$ where  $s_k = |E \left(\Theta_{F_k}^{*}\right)|$ refers to the total number of non-zero edges for the $k$th minipatch $F_k$. Also let $\varphi_{\max}()$ and $\varphi_{\min}()$ denote the maximum and minimum eigenvalue respectively. Denote the elementwise $\ell_{\infty}$-norm of a matrix by $\|\mathbf{A}\|_{\infty}=\max_{i\neq j}|A_{ij}|$.
 Consider the following assumptions:

\begin{assumption}[restate = assumptionMPGone, name = ]
\label{assumptionMPG1}
$X_{i} \text { be i.i.d. } \mathcal{N}\left(\mathbf{0}, \Sigma^* \right)$.
\end{assumption}
\begin{assumption}[restate = assumptionMPGtwo, name = ]
\label{assumptionMPG2}
 $\varphi_{\min }\left( \Sigma^* \right) \geq \underline{\kappa}>0,$ or equivalently $\varphi_{\max }\left(\bTheta^*\right) \leq 1 / \underline{\kappa}$.
\end{assumption}
\begin{assumption}[restate = assumptionMPGthree, name = ]
\label{assumptionMPG3}
$\varphi_{\max }\left(\Sigma^*\right) \leq \overline{\kappa}$.
\end{assumption}
\begin{assumption}[restate = assumptionMPGfour, name = ]
\label{assumptionMPG4}
Minimum edge strength:
\begin{equation*}
\theta_{\min }:=\min _{(i, j) \in E\left(\bTheta^{*}\right)}\left|\bTheta_{i j}^{*}\right| > c_1 \sqrt{\frac{s \log m}{n}}.
\end{equation*}
\end{assumption}
\hide{
\begin{assumption}[restate = assumptionMPGfiveraw, name = ]
$\max \limits_{F_k} \|  \left( \bTheta_{F_k}^{*}  -  \bTheta_{F_k F_k^c}^{*} \left(\bTheta_{F_k^c}^{*}\right)^{-1} \bTheta_{F_k^c F_k}^{*} \right)^{-1} - \left(\bTheta_{F_k}^{*}\right)^{-1}   \|_{\infty}  = \mathcal O\bigg( \sqrt{\frac{\log m}{n}} \bigg)$.
\end{assumption}
}
\begin{assumption}[restate = assumptionMPGfiveraw, name = ]

\label{assumptionMPG6}
Maximum effect of unsampled nodes:

\begin{align*}
 &\max \limits_{F_k} \|  \left( \bTheta_{F_k}^{*}  -  \bTheta_{F_k F_k^c}^{*} \left(\bTheta_{F_k^c}^{*}\right)^{-1} \bTheta_{F_k^c F_k}^{*} \right)^{-1} - \left(\bTheta_{F_k}^{*}\right)^{-1}   \|_{\infty} \\  &=
\mathcal{O}\left(\sqrt{\frac{\log m}{n}}\right)
\end{align*}
\end{assumption}

First, we have the following lemma on the boundedness of the eigenvalues of the covariance matrix of each minipatch.

\begin{lemma}
\label{MPGlemma1}
If Assumption~\ref{assumptionMPG2} and~\ref{assumptionMPG3} hold true, then $\varphi_{\min }\left( \Sigma_{F_k}^* \right) \geq \underline{\kappa}>0,$ and $\varphi_{\max }\left(\Sigma_{F_k}^*\right) \leq \overline{\kappa}$ for all minipatches $F_k$.
\end{lemma}

\noindent \underline{\textbf{Proof of Lemma~\ref{MPGlemma1}:} }
Denote minimal eigenvalue $\varphi_{\text {min}}$ and maximal eigenvalue $\varphi_{\max}$. Then we have
$$
\begin{aligned}
\varphi_{\text {min }}(\Sigma^*) &=\min_{v^T v = 1} v^T \Sigma^* v \\
\varphi_{\operatorname{max}}(\Sigma^*) &=\max_{v^T v = 1} v^T \Sigma^* v.
\end{aligned}
$$

\vspace{0.5em}
Note that the matrix $X_{F_k}^{T} X_{F_k}$ is a submatrix of $X^{T} X$. Namely, $X_{F_k}^{T} X_{F_k}$ consists of the rows and columns of $X^{T} X$ indexed by the elements of $F_k$. Define the projection:
$$
\widetilde{.}: \mathbb{R}^{p} \rightarrow \mathbb{R}^{m}, y=\left(y_{1}, \ldots, y_{p}\right)^{T} \mapsto \tilde{y}=\left(y_{F_{1}}, \ldots, y_{F_{k}}\right)^{T}.
$$
\vspace{0.5em}
Moreover, define the set of column vectors:
$$
V_{F_k}=\left\{x=\left(x_{1}, \ldots, x_{p}\right)^{T} \in \mathbb{R}^{p} \mid x^{T} x=1 \text { and } \forall j \notin F_k: x_{j}=0\right\} .
$$
\vspace{0.5em}
Note that
$$
\widetilde{V}_{F_k}=\left\{y \in \mathbb{R}^{m} \mid y^{T} y=1\right\}.
$$
\vspace{0.5em}
Furthermore, by construction, we have:
$$
\forall v \in V_{F_k}: v^{T} X^{T} X v=(\tilde{v})^{T} X_{F_k}^{T} X_{F_k} \tilde{v}.
$$
\vspace{0.5em}
Combining all these, we get
$$
\begin{aligned}
\varphi_{\min}(\Sigma^*_{F_k}) = \varphi_{\min}\left(X_{F_k}^{T} X_{F_k}\right) &=\min _{\left\{y \in \mathbb{R}^{m} \mid y^{T} y=1\right\}} y^{T} X_{F_k}^{T} X_{F_k} y \\
&=\min_{\tilde v \in \widetilde{V}_{F_k}}(\tilde{v})^{T} X_{F_k}^{T} X_{F_k} \tilde{v} \\
&=\min_{v \in V_{F_k}} v^{T} X^{T} X v \\
& \geq \min _{\left\{v \in \mathbb{R}^{p} \mid v^{T} v=1\right\}} v^{T} X^{T} X v \\
&=\varphi_{\min}\left(X^{T} X\right).
\end{aligned}
$$
\vspace{0.5em}
and similarly,
$$
\begin{aligned}
\varphi_{\max}(\Sigma^*_{F_k}) =  \varphi_{\max}\left(X_{F_k}^{T} X_{F_k}\right) &=\max _{\left\{y \in \mathbb{R}^{m} \mid y^{T} y=1\right\}} y^{T} X_{F_k}^{T} X_{F_k} y \\
&=\max _{v \in V_{F_k}}(\tilde{v})^{T} X_{F_k}^{T} X_{F_k} \tilde{v} \\
&=\max _{v \in V_{F_k}} v^{T} X^{T} X v \\
& \leq \max _{\left\{v \in \mathbb{R}^{p} \mid v^{r} v=1\right\}} v^{T} X^{T} X v \\
&=\varphi_{\max}\left(X^{T} X\right),
\end{aligned}
$$
as desired.

$\qed$


We now prove graph selection consistency by first proving that we have graph selection consistency for each minipatch.
\begin{lemma}
\label{MPGlemma4}
Let Assumptions~\ref{assumptionMPG1}-\ref{assumptionMPG6} be satisfied. For each minipatch $k$, denote the feature set $F_k$. Then for each minipatch $F_k$, the minipatch graph selection estimator with $\lambda \asymp \sqrt{\frac{\log m}{n}}$ is graph selection consistent:
\hide{
\begin{align*}
    &\mathbb{P}\left(\operatorname{sign}(\widetilde{\bTheta}^{(k)}_{ij})=\operatorname{sign}\left(\bTheta_{ij}^{*}\right), \forall \widetilde{\bTheta}^{(k)}_{ij} \in \widetilde{\bTheta}^{(k)}, i,j \in F_k \right)
    \to 1, \hspace{6mm} \forall i \neq j.
\end{align*}
}
\begin{align*}
    &\mathbb{P}\left(\operatorname{sign}(\widetilde{\bTheta}^{(k)}_{ij})=\operatorname{sign}\left(\bTheta_{ij}^{*}\right), \forall \widetilde{\bTheta}^{(k)}_{ij} \in \widetilde{\bTheta}^{(k)}, i,j \in F_k \right)  \\
     &\geq 1  - b_1 \exp(-b_2 n \lambda^2) \geq  1 - m^{\nu} \to 1, \hspace{6mm} \forall i \neq j.
\end{align*}
\end{lemma}

\noindent \underline{\textbf{Proof of Lemma~\ref{MPGlemma4}:} }
Lemma~\ref{MPGlemma1} suggests that, under Assumption~\ref{assumptionMPG2} and \ref{assumptionMPG3},
$\varphi_{\min }\left( \Sigma_{F_k}^* \right) \geq \underline{\kappa}>0,$ and $\varphi_{\max }\left(\Sigma_{F_k}^*\right) \leq \overline{\kappa}$ for all minipatches $F_k$.
\hide{
Lemma~\ref{MPGlemma3} suggests that, under Assumption~\ref{assumptionMPG5}, $ \|  (S_{F_k}^* - L_{F_k}^*)^{-1} -  (S_{F_k}^*)^{-1}   \|_{\infty}  = \mathcal O\bigg( \sqrt{\frac{\log p}{n}} \bigg)$ for all minipatches $F_k$. }

Note that the left term in Assumption~\ref{assumptionMPG6} is reminiscent of the quantity associated with the latent effects, $\eta$ in \cite{wang2021tglasso}, extended to the minipatches case.
We show that the thresholded graphical Lasso estimator is still graph selection consistent on each minipatch under our new assumption. Note that the quantity $\mathrm{I} $ in the proof of Proposition 2 of \cite{wang2021tglasso} now becomes:
\begin{align*}
\mathrm{I} &\leq \max_{i \neq j} \left|\hat{\sigma}_{i j}-\sigma_{s i j}\right| \cdot \left\|\Delta^{-}\right\|_{1} \\
&\leq  \max_{i \neq j} \left| (\hat{\sigma}_{i j}-\sigma_{0 i j}) + (\sigma_{0 i j} - \sigma_{s i j}) \right| \cdot \left\|\Delta^{-}\right\|_{1} \\
& \leq  \max_{i \neq j} (| \hat{\sigma}_{i j}-\sigma_{0 i j}  | + |  \sigma_{0 i j} - \sigma_{s i j} |) \cdot \left\|\Delta^{-}\right\|_{1} \\
& \leq 2 C_1 \sqrt{\frac{\log m}{n}}\left\|\Delta^{-}\right\|_{1},
\end{align*}
where the constant $C_1$ satisfies $ \max_{i \neq j} | \hat{\sigma}_{i j}-\sigma_{0 i j}  | \leq C_1 \sqrt{\frac{\log m}{n}}$ with probability tending to 1. The last inequality holds true due to Assumption~\ref{assumptionMPG6}. In particular, we assume
$\max \limits_{F_k} \|  \left( \bTheta_{F_k}^{*}  -  \bTheta_{F_k F_k^c}^{*} \left(\bTheta_{F_k^c}^{*}\right)^{-1} \bTheta_{F_k^c F_k}^{*} \right)^{-1} - \left(\bTheta_{F_k}^{*}\right)^{-1}   \|_{\infty}  \leq C_1 \sqrt{\frac{\log m}{n}} $.
Here, in our setup, $\hat \sigma_{ij}$ refers to the entries of $\hat \Sigma_{F_k}$, $\sigma_{0ij}$ refers to the entries of $\Sigma^*_{F_k}$, $ \sigma_{s i j}$ refers to  the entries of $(\Theta^*_{F_k})^{-1}$ and $\Delta = \widehat \Theta_{F_k} - \Theta^*_{F_k}$.
The rest of the proof follows as \cite{wang2021tglasso}  paper. Therefore, Assumption~\ref{assumptionMPG6} still guarantees that the thresholded graphical Lasso estimator is graph selection consistent  for the $k$th minipatch $F_k$.

\vspace{0.5em}
Hence, all the assumptions required for Theorem 3 in the work of \cite{wang2021tglasso} are satisfied. Therefore, we have graph selection consistency for all minipatches $F_k$, i.e.,
\hide{
\begin{align*}
    &\mathbb{P}\left(\operatorname{sign}(\widetilde{\bTheta}^{(k)}_{ij})=\operatorname{sign}\left(\bTheta_{ij}^{*}\right), \forall \widetilde{\bTheta}^{(k)}_{ij} \in \widetilde{\bTheta}^{(k)}, i,j \in F_k \right)
    \to 1, \hspace{6mm} \forall i \neq j.
\end{align*}}
\begin{align*}
    &\mathbb{P}\left(\operatorname{sign}(\widetilde{\bTheta}^{(k)}_{ij})=\operatorname{sign}\left(\bTheta_{ij}^{*}\right), \forall \widetilde{\bTheta}^{(k)}_{ij} \in \widetilde{\bTheta}^{(k)}, i,j \in F_k \right)  \\
     &\geq 1  - b_1 \exp(-b_2 n \lambda^2) \geq  1 - m^{\nu} \to 1, \hspace{6mm} \forall i \neq j.
\end{align*}
for all minipatches $F_k$. Here, $b_1$ and $b_2$ depend on $\overline{\kappa}$ only, as in Lemma 1 of \cite{rothman2008sparse} and Lemma 3 of \cite{bickel2008regularized}.

$\qed$

Now, let $R_{ij}$ denote the number of times any pair of nodes $i,j$ are sampled together into minipatches.  Define constant $\nu$ (as in Lemma~\ref{MPGlemma4}) to satisfy $m^{\nu} = b_1 \exp(-b_2 n \lambda^2)$ where $b_1$ and $b_2$ depend on $\overline{\kappa}$; 
additionally, define $b_0 = \min \left\{ \frac{(1-m^{\nu} - \pi_{\text{thr}})^2 }{2},   \frac{( \pi_{\text{thr}}-m^{\nu} )^2 }{2} \right\}$; finally, define the constant $L$ that satisfies $|S_{ij}^{(k)}  - (1-m^{\nu}) | \leq L$ for all pairs of $(i,j) \in E^*$ that are sampled on the $k$th minipatch and $|S_{ij}^{(k)}  - m^{\nu} | \leq L$ for all pairs of $(i,j) \in {E^*}^c$ that are sampled on the $k$th minipatch.
Given these, we have the following theorem on the graphical model selection consistency of our MPGraph approach.

\begin{theorem}
\label{MPGtheorem}
Let Assumptions~\ref{assumptionMPG1}-\ref{assumptionMPG6}  be satisfied and let $n$ grow proportionally with $N$. Then, the minipatch graph selection estimator, MPGraph, with $\lambda \asymp \sqrt{\frac{\log m}{n}}$ and $\tau \asymp \sqrt{\frac{s \log m}{n}}$, is graph selection consistent with high probability:

\begin{align*}
    & \mathbb{P}\left(  \hat{E}^{\text{stable}} =  E^*\right)   \\
    &\geq  1 -   \exp\left\{ - \min R_{ij} \cdot \frac{(  (1-m^{\nu}) - \pi_{\text{thr}}  )^2/2}{  m^{\nu} + L/3 }  +  \textcolor{black}{2 \ln m}      \right\} - \\
    &\exp\left\{ - \min R_{ij} \cdot \frac{(  \pi_{\text{thr}} - m^{\nu}   )^2/2}{  m^{\nu} + L/3 }  + \textcolor{black}{2 \ln m}     \right\}   \\
    & \geq 1 - 2 \exp \left\{ - b_0 \cdot \frac{\min_{i,j} R_{ij}}{m^{\nu} + L  / 3 }  + \textcolor{black}{2 \ln m}  \right\} \\
    & \to 1 ,\hspace{6mm} \text{ as } N \to \infty,  \text{ or as } K\to \infty.
\end{align*}
\hide{where $R_{ij}$ is the number of times any pair of nodes $i,j$ are sampled together into minipatches;
\textcolor{red}{$\nu$ refers to the constant in Lemma~\ref{MPGlemma4} with $m^\nu  = b_1 \exp(-b_2 n \lambda^2)$ where $b_1$ and $b_2$ depends on $\overline{\kappa}$ and $\underline{\kappa}$ .}}
\end{theorem}

\noindent \underline{\textbf{Proof of Theorem~\ref{MPGtheorem}:} }
By definition, $\hat{E}^{\text{stable}} = \big\{1\leq i < j \leq M: \hat{\Pi}^{(K)}_{ij} \geq \pi_{\text{thr}} \big\}$. Meanwhile,
\begin{align*}
    \mathbb P (\hat{\Pi}^{(K)}_{ij} \geq \pi_{\text{thr}}) &= \mathbb P\left( \frac{\sum_{k=1}^K S_{ij}^{(k)}}{\text{max}(1,\sum_{k=1}^K D_{ij}^{(k)})} \geq \pi_{\text{thr}}\right)  \\
    & = \mathbb P\left( \sum_{k=1}^K S_{ij}^{(k)} \geq \pi_{\text{thr}} \cdot {\text{max}(1,\sum_{k=1}^K D_{ij}^{(k)})} \right) .
\end{align*}

We first consider the case of true edges, $(i,j) \in E^*$; the case for null edges will be discussed later. Denote $R_{ij} = \sum_{k=1}^K D_{ij}^{(k)}$ as the number of times both nodes $i$ and $j$ are sampled together into minipatches. Without loss of generality, we assume that each pair of nodes is sampled at least once  during the  MPGraph procedure.

\vspace{0.5em}
From Lemma~\ref{MPGlemma4}, we know that, for a true edge $(i,j) \in E^*$, $S_{ij}^{(k)}$ is a Bernoulli random variable with success probability at least $1 - m^{\nu}$ given that the pair $(i,j)$ are sampled on the $k$th minipatch, where $\nu$ depends on $n$, $\overline{\kappa}$ and $\underline{\kappa}$; hence, $\sum_{k=1}^K  S_{ij}^{(k)} \overset{i.i.d.}{\sim}$ Binomial$(R_{ij},1-m^{\nu})$. Here, without loss of generality, we consider the case when the lower bound of Lemma~\ref{MPGlemma4} is attained, i.e., success probability equals $1 - m^{\nu}$; results when the success probability is greater than $ 1 - m^{\nu}$ still hold.

\vspace{0.5em}
To calculate the probability above, we use the Bernstein's inequality for bounded distributions. Let $X_{1}, \ldots, X_{n}$ be independent, mean zero random variables, such that $|X_i| \leq L$ with probability one, for all $i$. Then, for every $t \geq 0$, we have
\begin{align*}
\mathbb{P}\left(\left|\sum_{i=1}^{n} X_{i}\right| \geq t\right) \leq 2 \exp \left(-\frac{t^{2} / 2}{\sigma^{2}+L t / 3}\right) ,
\end{align*}
where $\sigma^{2}=\sum_{i=1}^{n} \mathbb{E} X_{i}^{2}$ is the variance of the sum.

\vspace{0.5em}
Here, in our case, we have
\begin{align*}
    &\mathbb P (\hat{\Pi}^{(K)}_{ij} \geq \pi_{\text{thr}})  \\
    & =  \mathbb P\left( \sum_{k=1}^K S_{ij}^{(k)} \geq \pi_{\text{thr}} R_{ij}  \right)   =  1 -   \mathbb P\left( \sum_{k=1}^K S_{ij}^{(k)} < \pi_{\text{thr}} R_{ij}  \right) \\
    & = 1 - \mathbb P\left( \sum_{k=1}^K S_{ij}^{(k)} - R_{ij}(1-m^{\nu}) < \pi_{\text{thr}} R_{ij}  - R_{ij}(1-m^{\nu})  \right)  \\
    & = 1 - \mathbb P\left( \sum_{k=1}^K S_{ij}^{(k)} - R_{ij}(1-m^{\nu}) < - R_{ij} (1-m^{\nu}- \pi_{\text{thr}})  \right)  \\
    & \geq 1 - \exp\left\{ - \frac{( R_{ij}(1-m^{\nu} - \pi_{\text{thr}}))^2/2}{R_{ij} m^{\nu} (1-m^{\nu}) + L R_{ij} (1-m^{\nu}- \pi_{\text{thr}})/3 }        \right\}  \\
    & \geq 1 - \exp\left\{ -R_{ij} \frac{( (1-m^{\nu} - \pi_{\text{thr}}))^2/2}{ m^{\nu}  + L  /3 }        \right\} \\
    & = 1 - \exp\left\{ - R_{ij}\frac{(  (1-m^{\nu}) - \pi_{\text{thr}}  )^2/2}{  m^{\nu}  }        \right\}, \quad  \text{ as } m \to \infty.
\end{align*}
\noindent Or equivalently, $\mathbb P (\hat{\Pi}^{(K)}_{ij} \leq \pi_{\text{thr}}) \leq \exp\left\{ - R_{ij}\frac{(  (1-m^{\nu}) - \pi_{\text{thr}}  )^2/2}{  m^{\nu}  }        \right\}$ as $m \to \infty$.

Here, the tuning parameter $\pi_{\text{thr}}$ is chosen such that $1 - m^{\nu} \geq \pi_{\text{thr}}$.
The last inequality holds true as $1-m^{\nu} \leq 1$ and $1-m^{\nu}- \pi_{\text{thr}} \leq 1$.
The constant $L$ satisfies $|S_{ij}^{(k)}  - (1-m^{\nu}) | \leq L$ for all pairs of $(i,j)$ that are sampled on the $k$th minipatch. When $m \to \infty$, $1-m^{\nu} \to 1$. Therefore, the last equality holds true as $S_{ij}^{(k)} = 1$ and thus $L = 0$ with probability to 1.

By the union bound, for all true edges, we have:
\begin{align*}
    & \mathbb P \left(\hat{\Pi}^{(K)}_{ij} \geq \pi_{\text{thr}} , \forall (i,j) \in E^*\right)  \\
    &=    1 - \mathbb P \left(\hat{\Pi}^{(K)}_{ij} \leq \pi_{\text{thr}} , \exists (i,j) \in E^*\right)  \\
    & \geq  1 - m^2 \exp\left\{ - \frac{( R_{ij}(1-m^{\nu} - \pi_{\text{thr}}))^2/2}{R_{ij} m^{\nu} (1-m^{\nu}) + L R_{ij} (1-m^{\nu}- \pi_{\text{thr}})/3 }        \right\}  \\
    &  \geq 1 - m^2 \exp\left\{ - \min R_{ij} \cdot \frac{(  (1-m^{\nu}) - \pi_{\text{thr}}  )^2/2}{ m^{\nu} (1-m^{\nu}) + L   (1-m^{\nu}- \pi_{\text{thr}})/3  }        \right\}    \\
    & \geq 1 - m^2 \exp\left\{ - \min R_{ij} \cdot \frac{(  (1-m^{\nu}) - \pi_{\text{thr}}  )^2/2}{ m^{\nu} + L   /3  }        \right\}    \\
    & = 1 - \exp\left\{ - \min R_{ij} \cdot \frac{(  (1-m^{\nu}) - \pi_{\text{thr}}  )^2/2}{  m^{\nu} }  + \textcolor{black}{2 \ln m}      \right\} , \\  &\text{ as } m \to \infty.
\end{align*}
\vspace{0.5em}
Note $\min R_{ij}$ refers to the minimum number of times any pair of nodes are sampled together into  minipatches. We know that when $m \to \infty$, $   m^{\nu} (1-m^{\nu}) \to 0 $ for $\nu < 0$. \textcolor{black}{Further, when $\nu < -1$, we have $m^{-\nu} > \ln m$ and $- m^{-\nu} + \ln m \to -\infty $.} Therefore, $ -  \min R_{ij} \cdot \frac{(  (1-m^{\nu}) - \pi_{\text{thr}}  )^2/2}{  m^{\nu} } + 2\ln m \to - \infty $ when $m \to \infty$ or $R_{ij} \to \infty$, and
we have $\mathbb P (\hat{\Pi}^{(K)}_{ij} \geq \pi_{\text{thr}}  , \forall (i,j) \in E^* )   \to 1$.

\vspace{0.5em}
Similarly, for a null edge $(i,j) \in {E^*}^{c}$, $S_{ij}^{(k)}$ is a Bernoulli random variable with success probability $m^{\nu}$; hence, $\sum_{k=1}^K  S_{ij}^{(k)} \overset{i.i.d.}{\sim}$ Binomial$(R_{ij}, m^{\nu})$.

\vspace{0.5em}

Again, applying the Bernstein's inequality, we have
\begin{align*}
    & \mathbb P (\hat{\Pi}^{(K)}_{ij} <  \pi_{\text{thr}})  \\
    &=  \mathbb P\left( \sum_{k=1}^K S_{ij}^{(k)}  < \pi_{\text{thr}} R_{ij}  \right)   =  1 -   \mathbb P\left( \sum_{k=1}^K S_{ij}^{(k)} \geq \pi_{\text{thr}} R_{ij}  \right) \\
    & = 1 - \mathbb P\left( \sum_{k=1}^K S_{ij}^{(k)} - R_{ij}  m^{\nu}  \geq \pi_{\text{thr}} R_{ij}  - R_{ij} m^{\nu}   \right)  \\
    & = 1 - \mathbb P\left( \sum_{k=1}^K S_{ij}^{(k)} - R_{ij}  m^{\nu}  \geq R_{ij}  (\pi_{\text{thr}}  - m^{\nu})   \right)  \\
    & \geq 1 - \exp\left\{ - \frac{(   R_{ij} (\pi_{\text{thr}}  -  m^{\nu}))^2/2 }{R_{ij} m^{\nu} (1-m^{\nu}) + L R_{ij}  (\pi_{\text{thr}}  - m^{\nu}) /3 }        \right\}  \\
    & \geq 1 - \exp\left\{ - R_{ij} \frac{   (\pi_{\text{thr}}  -  m^{\nu})^2/2 }{ m^{\nu} + L /3 }        \right\}  \\
    & = 1 - \exp\left\{ - R_{ij}\frac{(  \pi_{\text{thr}} - m^{\nu}  )^2/2}{  m^{\nu}  }        \right\}, \quad  \text{ as } m \to \infty.
\end{align*}

Here, the tuning parameter $\pi_{\text{thr}}$ is chosen such that $\pi_{\text{thr}}   \geq m^{\nu}$. Still, the constant $L$ satisfies $|S_{ij}^{(k)}  - m^{\nu} | \leq L$ for all pairs of $(i,j)$ that are sampled on the $k$th minipatch. When $m \to \infty$, $m^{\nu} \to 0$. Therefore, the last equality holds true as $S_{ij}^{(k)} = 0$ and thus $L = 0$ with probability to 1. By the union bound, for all null edges, we have:
\begin{align*}
    & \mathbb P \left(\hat{\Pi}^{(K)}_{ij} <  \pi_{\text{thr}}, \forall (i,j) \in {E^*}^{c}\right)  \\
    &=    1 - \mathbb P \left(\hat{\Pi}^{(K)}_{ij} \geq \pi_{\text{thr}} , \exists (i,j) \in {E^*}^{c}\right)  \\
    & \geq  1 - m^2 \exp\left\{ - \frac{(   R_{ij} (\pi_{\text{thr}}  -  m^{\nu}))^2/2 }{R_{ij} m^{\nu} (1-m^{\nu}) + L R_{ij}  (\pi_{\text{thr}}  - m^{\nu}) /3 }        \right\}   \\
    &  \geq 1 - m^2 \exp\left\{ - \min R_{ij} \cdot \frac{(   \pi_{\text{thr}} - m^{\nu}   )^2/2}{  m^{\nu} (1-m^{\nu}) + L   (\pi_{\text{thr}}  - m^{\nu}) /3 }        \right\}  \\
    &  \geq 1 - m^2 \exp\left\{ - \min R_{ij} \cdot \frac{(   \pi_{\text{thr}} - m^{\nu}   )^2/2}{  m^{\nu} + L    /3 }        \right\}  \\
    & = 1 - \exp\left\{ - \min R_{ij} \cdot \frac{(  \pi_{\text{thr}} - m^{\nu}   )^2/2}{ m^{\nu}  }  + \textcolor{black}{2 \ln m}     \right\}, \quad  \text{ as } m \to \infty.
\end{align*}

\vspace{0.5em}
Similarly, we have $\mathbb P (\hat{\Pi}^{(K)}_{ij} <  \pi_{\text{thr}}, \forall (i,j) \in {E^*}^{c})  \to 1$ when $m \to \infty$ or $R_{ij} \to \infty$.

\vspace{0.5em}
Combining the two cases, we have:
\begin{align*}
    &\mathbb{P}\left(  \hat{E}^{\text{stable}} =  E^*\right)  \\
    & = 1 - P \left(\hat{\Pi}^{(K)}_{ij}  < \pi_{\text{thr}} , \exists (i,j) \in E^* \text{ or } \hat{\Pi}^{(K)}_{ij} \geq  \pi_{\text{thr}}, \exists (i,j) \in {E^*}^{c} \right)  \\
    & \geq 1 -   \exp\left\{ - \min R_{ij} \cdot \frac{(  (1-m^{\nu}) - \pi_{\text{thr}}  )^2/2}{  m^{\nu} (1-m^{\nu}) +  L   (1-m^{\nu}- \pi_{\text{thr}})/3 }  + \textcolor{black}{2 \ln m}      \right\} - \\
    &  \exp\left\{ - \min R_{ij} \cdot \frac{(  \pi_{\text{thr}} - m^{\nu}   )^2/2}{  m^{\nu} (1-m^{\nu}) + L   (\pi_{\text{thr}}  - m^{\nu}) /3}  + \textcolor{black}{2 \ln m}     \right\}  \\
    & \geq 1 -   \exp\left\{ - \min R_{ij} \cdot \frac{(  (1-m^{\nu}) - \pi_{\text{thr}}  )^2/2}{  m^{\nu} +  L   /3 }  + \textcolor{black}{2 \ln m}      \right\} - \\
    &  \exp\left\{ - \min R_{ij} \cdot \frac{(  \pi_{\text{thr}} - m^{\nu}   )^2/2}{  m^{\nu}  + L    /3}  + \textcolor{black}{2 \ln m}     \right\}  \\
    & \geq 1 - 2 \exp \left\{ - b_0 \cdot \frac{\min_{i,j} R_{ij}}{m^{\nu} + L  / 3 } + \textcolor{black}{2 \ln m}  \right\} \\
    & \to 1 , \hspace{6mm}  \text{ as } N \to \infty,  \text{ or as }
    K \to \infty.
\end{align*}
where $b_0 = \min \left\{ \frac{(1-m^{\nu} - \pi_{\text{thr}})^2 }{2},   \frac{( \pi_{\text{thr}}-m^{\nu} )^2 }{2} \right\}$.

$\qed$



\section{Additional Empirical Studies to Investigate Effects of Minipatch Size}
\label{sec:mpgraph-minipatch-size}

\begin{figure}[!ht]
    \centering
    \includegraphics[width=0.7\linewidth]{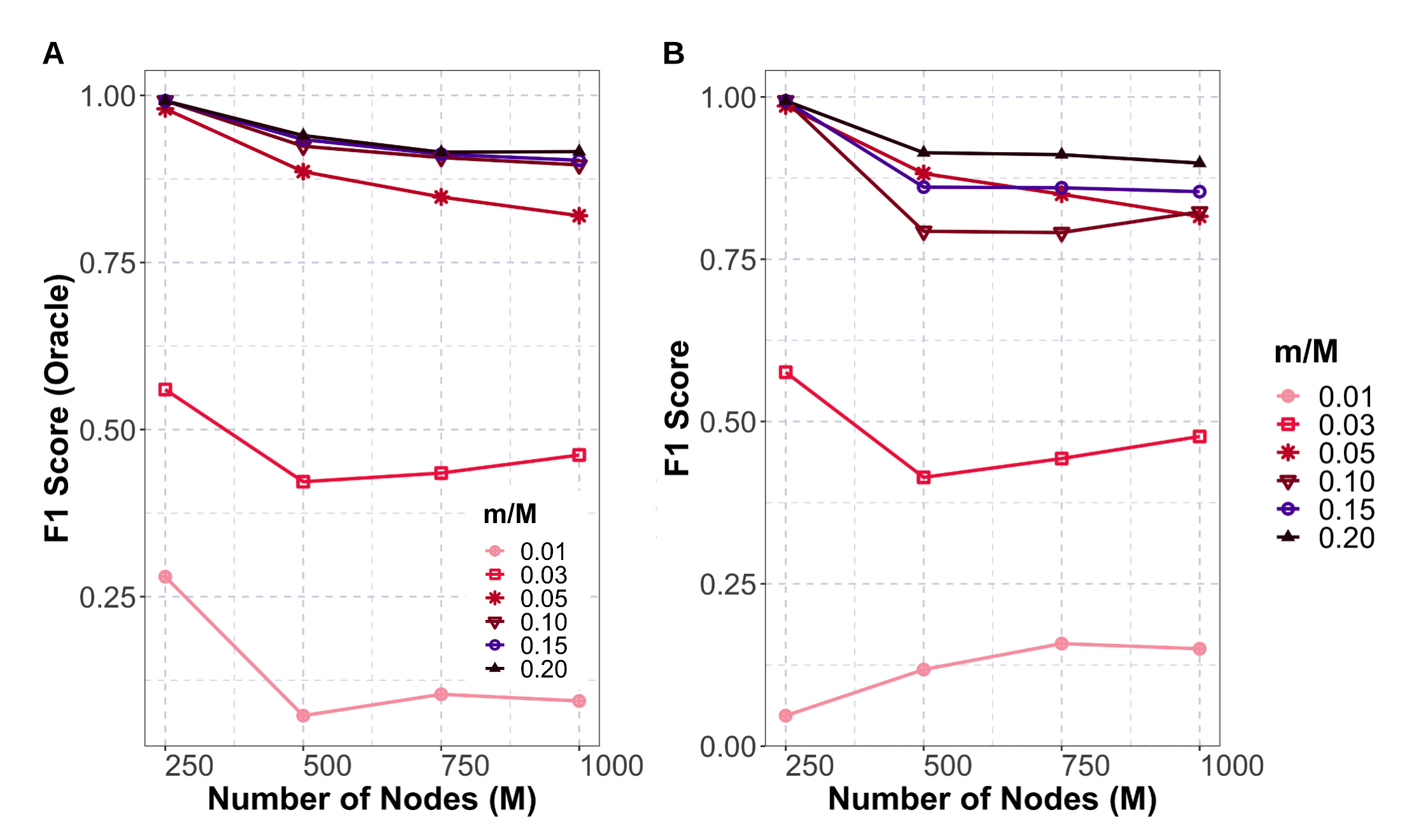}
    \caption{Performance of MPGraph for Small-World Graph Simulations with Fixed $N=500$ and Varying Dimensionality $M$. (A) Edge selection accuracy of MPGraph using various minipatch sizes (i.e. $m/M$) with oracle tuning approaches (i.e. assume total number of true edges $|E|$ is known). (B) Edge selection accuracy of MPGraph using various minipatch sizes (i.e. $m/M$) for the same simulations in (A), but with data-driven tuning approaches.}
    \label{minipatchsize}
    \end{figure}

In Figure~\ref{minipatchsize}A, we empirically demonstrate that the number of nodes in a minipatch $m$ is the determining factor for how well MPGraph recovers the true graph. For the challenging small-world graph situations in Figure~\ref{minipatchsize}A, the performance of MPGraph degrades precipitously when $m$ is too small. But with a sufficiently large $m$ (e.g. at least $5\%$ of $M$), MPGraph achieves strong performance. Because the total number of true edges $|E|$ is usually unknown in realistic use cases of graphical model selection in practice, we also show the edge selection accuracy of MPGraph with data-driven tuning (i.e. $\hat{E}^{\text{stable}} = \big\{1\leq i < j \leq M: \hat{\Pi}^{(K)}_{ij} \geq \pi_{\text{thr}} \big\}$ with $\pi_{\text{thr}}=0.5$) in Figure~\ref{minipatchsize}B. As we can see, the performance of MPGraph with data-driven tuning is close to that of MPGraph with oracle tuning.

\begin{figure}[!ht]
    \centering
    \includegraphics[width=0.7\linewidth]{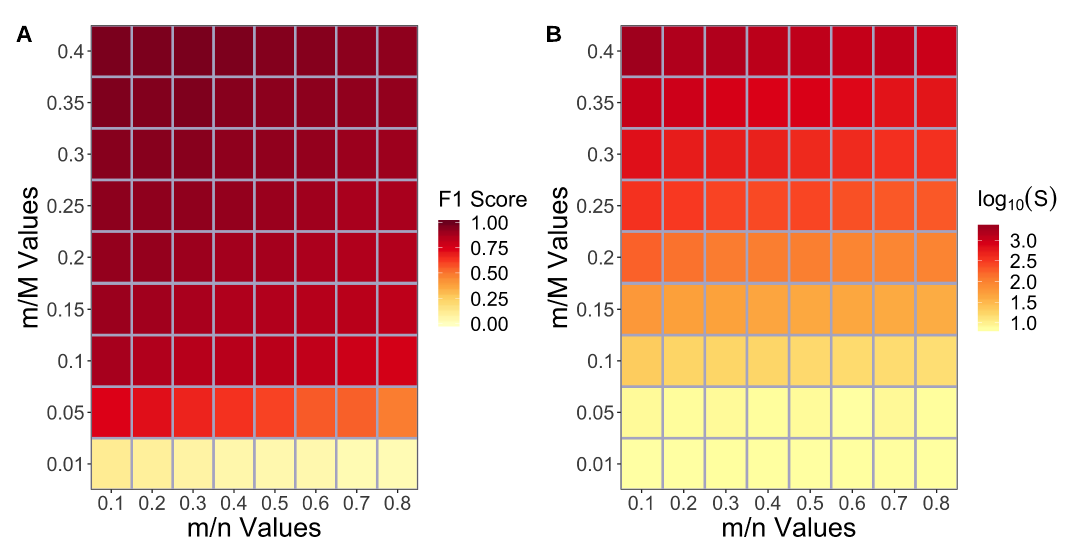}
    \caption{Effects of Minipatch Size. We demonstrate how edge selection accuracy and computational time of MPGraph change with different minipatch sizes in terms of $m/M$ and $m/n$, where $M$ is the total number of nodes. (A) Edge selection accuracy in terms of F1 Score (with data-driven tuning); (B) Computational time on $\log_{10}(\text{second})$ scale. We see that our method has stable edge selection accuracy for a sensible range of $n$ and $m$ values.}
    \label{heatmap}
    \end{figure}

\vspace{0.5em}
We have more empirical studies investigating how minipatch sizes impact on edge selection accuracy and computational time of MPGraph. Following the same simulation procedure outlined in Sec. III of the main paper, we generate data $\mathbf{X}\in\mathbb{R}^{3000\times 500}$ from the groundtruth precision matrix that has the Erd\H{o}s-Renyi graph structure. To investigate the effects of minipatch size on performance, we run MPGraph on a grid of $n$ and $m$ values. In particular, we use a sequence of $m$ values such that $m/M\in\{0.01, 0.05, \hdots,0.4\}$ and then choose $n$ relative to $m$ so that $m/n \in \{0.1, \hdots, 0.8\}$. All other parameters are set to the default values (i.e. $\pi_{\text{thr}}=0.5$ and $K=1000$). Edge selection accuracy in terms of the F1 Scores and computational time are reported for these minipatch sizes in Figure \ref{heatmap}. We see that MPGraph has stable edge selection accuracy for a sensible range of $n$ and $m$ values. As a general rule of thumb, we recommend taking $m$ to be $5\%\sim 10\%$ of $M$ and then picking $n$ relative to $m$ so that it surpasses the sample complexity of the base thresholded graph estimator used on the minipatches, which would strike a good balance between statistical accuracy and computational time in practice.

\end{document}